\pdfoutput=1

\documentclass[11pt]{article}

\usepackage[]{EMNLP2023}

\usepackage{times}
\usepackage{latexsym}

\usepackage[T1]{fontenc}

\usepackage[utf8]{inputenc}

\usepackage{microtype}

\usepackage{inconsolata}

\usepackage{pdflscape}
\usepackage{enumitem}
\usepackage{xcolor}
\usepackage{amsfonts}
\usepackage{amsmath}
\usepackage{bbm}
\usepackage{amsthm}
\usepackage{amssymb}
\usepackage{times}
\usepackage{here}
\usepackage{latexsym}
\usepackage{booktabs}
\usepackage{svg}

\usepackage{graphicx, natbib}

\usepackage{subfig}
\usepackage{cleveref}
\usepackage{cancel}
\usepackage{macro}

\crefname{section}{\S}{\S\S}
\crefname{table}{Tab.}{}
\crefname{figure}{Fig.}{}
\crefname{algorithm}{Algorithm}{}
\crefname{equation}{eq.}{}
\crefname{appendix}{App.}{}
\crefname{thm}{Theorem}{}
\crefname{prop}{Proposition}{}
\crefname{cor}{Corollary}{}
\crefname{observation}{Observation}{}
\crefname{assumption}{Assumption}{}
\crefformat{section}{\S#2#1#3}

\usepackage{comment}
\usepackage[disable]{todonotes}

\newcommand{\note}[4][]{\todo[author=#2,color=#3,size=\scriptsize,fancyline,caption={},#1]{#4}} 
\newcommand{\ryan}[2][]{\note[#1]{ryan}{violet!40}{#2}}

\newcommand{\response}[1]{\vspace{3pt}\hrule\vspace{3pt}\textbf{#1:}} 

\newcommand{\mathcheck}[1]{{\textcolor{black}{#1}}}
\usepackage{ifthen}
\usepackage{xargs}
\newcommandx*{\mathchecknew}[5][1,2,3,4,usedefault]{%
\ifthenelse { \equal {#1} {} }%
{{\color{BrickRed} #5}}%
{%
    \ifthenelse { \equal {#2} {} }%
    {{\color{VioletRed} #5}}
    {%
        \ifthenelse { \equal {#3} {} }%
        {{\color{YellowOrange} #5}}
        {%
            \ifthenelse { \equal {#4} {} }%
            {{\color{Aquamarine} #5}}
            {%
                {\color{ForestGreen} #5}
            }%
        }%
    }%
}%
}

\title{Towards Explainability in Legal Outcome Prediction Models}

\author{
  Josef Valvoda
  \\
  University of Cambridge
  \\
  \texttt{\href{mailto:jv406@cam.ac.uk}{jv406@cam.ac.uk}}
  \\
  \And
  Ryan Cotterell 
  \\
  ETH Z\"urich
  \\
  \texttt{\href{mailto:ryan.cotterell@inf.ethz.ch} {ryan.cotterell@inf.ethz.ch}} 
}

\date{}

\begin{document}
\maketitle
\begin{abstract}
    Current legal outcome prediction models---a staple of legal NLP---do not explain their reasoning.
    However, to employ these models in the real world, human legal actors need to be able to understand the model's decisions.
    In the case of common law, legal practitioners reason towards the outcome of a case by referring to past case law, known as precedent.
    We contend that precedent is, therefore, a natural way of facilitating explainability for legal NLP models.
    In this paper, we contribute a novel method for identifying the precedent employed by legal outcome prediction models.
    Furthermore, by developing a taxonomy of legal precedent, we are able to compare human judges and neural models with respect to the different types of precedent they rely on.
    We find that while the models learn to predict outcomes reasonably well, their use of precedent is unlike that of human judges.\looseness-1

\vspace{0.5em}
\hspace{.5em}\includegraphics[width=1.25em,height=1.25em]{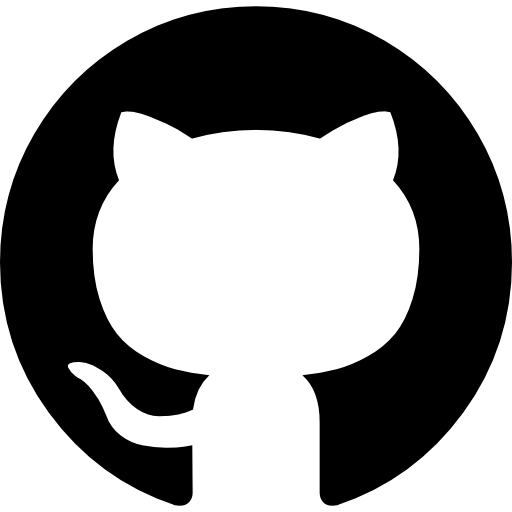}\hspace{.75em}
\parbox{\dimexpr\linewidth-7\fboxsep-7\fboxrule}{\url{https://github.com/valvoda/under_the_influence}}
\vspace{-.5em}
\end{abstract}
    
\section{Introduction}\label{sec:intro}
Legal outcome prediction models have generated much interest in the past years \cite{aletras, masha, t-y-s-s-etal-2023-zero, valvoda-etal-2023-role}.
Given a text transcript of the facts of a legal case, a legal outcome prediction model predicts whether a particular law has been breached.
Researchers have evaluated these models in several jurisdictions, including the UK, USA, Switzerland, France, China, Japan, India and the European Court of Human Rights \citep[ECtHR;][]{sulea-etal-2017-predicting, chalkidis-etal-2019-neural, ukcourt, malik-etal-2021-ildc, niklaus-etal-2021-swiss, chalkidis-etal-2022-lexglue, feng-etal-2022-legal}.

While legal outcome prediction models are interesting in the context of academic research \cite{valvoda-etal-2021-precedent}, any real-world deployment of Legal AI necessitates a certain level of understanding of how the system works \cite{valvoda2023ethics}.
After all, it is the \emph{reasoning} towards the outcome of the case, rather than the outcome itself, that is of tantamount importance in legal outcome prediction.
There are two reasons for this. 
(1) To be able to use a legal outcome prediction model system in practice implicitly assumes explainability: lawyers are bound to provide advice that the client can depend on. 
This obligation is so extreme that a negligent lawyer themself can be sued for misconduct if they give faulty advice.
Moreover, making legal decisions without understanding the reasoning behind them is unethical in many legal systems.
For example, for a model to be used in practice it is a legal requirement in the EU under GDPR to know what the decision is based on.
Thus, the ability to double-check any information a hypothetical legal AI assistant provides is paramount when building technology for legal professionals. 
(2) Explainability can create new use cases for legal outcome prediction models.
Knowing how to reason towards a desired legal outcome is valuable for drafting legal arguments and, thus, a legal outcome prediction system could be deployed as an assistive technology in areas well beyond simply predicting outcomes \cite{sarker}.\looseness-1

State-of-the-art legal outcome prediction models are nearly homogeneous at a technical level---they are all parameterized as neural network classifiers on top of pre-trained language models (LM).
While their performance is remarkable at times, the reasoning embedded in the model is largely inscrutable.
Despite the widespread recognition of the problem that unexplainable legal NLP systems pose \cite{explainablelegalAI} and emerging work in the domain of case law summarization \cite{norkute_explainable}, much remains to be done in order to build effective explainable legal NLP systems \cite{brantingexplainable, chalkidis-etal-2021-paragraph}.
In particular, the existing work on explainable legal AI focuses on providing explanations by identifying relevant tokens \cite{xu-etal-2023-dissonance}, sentences \cite{malik-etal-2021-ildc} or paragraphs \cite{chalkidis-etal-2021-paragraph}, from the facts of the case.
However, lawyers and judges typically reason at the level of a case.\looseness=-1

In this paper, we address the lack of explainability of legal outcome prediction models in a manner inspired by legal reasoning.
The law in the Anglo-American tradition is endowed with a natural rationale---precedent---which produces a legal argument.
Precedent is how judges and lawyers reason with respect to previous cases.
Because precedent is binding, by citing previously decided cases, legal actors substantiate their arguments.
Unlike existing explainability methods,
our method produces an explanation in terms of legal precedent.
This allows legal practitioners to easily inspect the legal reasoning the model employs.
Furthermore, we develop a taxonomy of legal precedent, which allows us to categorize each training case in terms of its precedential operation.
We can therefore pinpoint which kinds of precedential reasoning the models rely on.\looseness-1

We validate our proposed method by training several legal outcome prediction models and measuring the correlation between the precedent they use and that used by a human judge.
Overall, our experiments reveal that the precedent used by a human judge has a weak positive correlation with the precedent our models rely on (the highest Spearman's $\rho$ we observe is $0.18$).
We also discover that the model predictions positively correlate with one type of precedent in particular---the one where the outcome of the precedent case is the same as the outcome of the case at hand.
However, for most of the categories of precedent, we find the models have a negligible or even negative correlation with the precedent applied by a human judge.
Worse yet, model $F_1$ performance on predicting the outcome of a case is a poor proxy for its alignment with precedential reasoning; see \cref{table:results}.
In other words, higher $F_1$ does not equal better alignment.
Therefore, our results cast doubt on the real-world utility of the current outcome prediction models.\looseness-1

\section{Precedent-based Interpretability}\label{sec:motivation}

We now give an argument for why precedent-based interpretability is the right direction for improving the usability of legal outcome prediction models.\looseness=-1

\paragraph{Why are existing methods insufficient?}
Many interpretability methods, such as linguistic probing, have been developed to investigate the linguistic capabilities of neural models \cite{hewitt-manning-2019-structural, belinkov, pimentel-etal-2022-attentional}.
These methods often focus solely on the immediate input to the model.
This is intentional because the goal of linguistic probing is to analyze the model in terms of how individual tokens affect the output.
For example, if you want to know if and how a neural network's representations relate to a dependency parse of the input sentence, investigating the network's use of tokens in a sentence is a wise choice.
In legal NLP, however, despite the recent rejuvenation of efforts in the field \cite{xu-etal-2023-dissonance, habernal}, there is no generally accepted legal theory about how individual words or sentences affect the outcome of a case.
We further elaborate on the related work in \cref{sec:related}.\looseness-1

\paragraph{Why precedent?}
There is already a generally accepted method of legal reasoning---reasoning through legal precedent.
However, precedent does not operate on the granularity of individual words or sentences.
Instead, precedent relies on the relationship between the case at hand and previously decided cases.
Importantly, precedent is a way of developing law.
A decision rendered in a past case binds future decisions in a common law system.
This binding precedent, which stems from the doctrine of \textit{stare decisis}, literally \textit{stand by the decision}, ensures the reliability of law by forcing judges to make consistent decisions \citep{duxbury_2008, sep-legal-reas-prec, garner2009black}.\footnote{There is also a non-binding precedent. It is typically contained in \textit{obiter dicta} or the \textit{other things said} in the court decisions. Its purpose is to hint at what the court might decide in the future given a legal point that falls outside the scope of the case under consideration. 
Another type of non-binding precedent is in cases that have been overruled by a subsequent judgment. 
In such instances, the case is considered outdated and the law contained within is obsolete. While this kind of precedent is important, we do not study them in this paper.}
While precedent plays an important role in deciding the outcome of a case in both common and civil law systems, it occupies the core of legal reasoning in common law jurisdictions, e.g., UK, USA, and India, and it is the focus of this paper.\looseness=-1

\paragraph{Claims and Outcomes.}
Because precedent is binding law, it is used in legal argumentation.
When a new legal case arises, there are two competing points of view presented to a judge:
That of the plaintiff and that of the defendant.
These are typically represented by lawyers working on behalf of either side.
The first step in the process is establishing the \textbf{claims}, i.e., the laws that the plaintiff claims as violated. 
The claims are a starting point of the legal argument. 
The job of the lawyer on either side is to substantiate their claims.
To do this, the lawyers rely on previous case law, i.e., the precedent.
They will draw analogies to cases where their side has won, and distinguish their circumstances from the instances where their side has lost.
The selection of past cases is therefore a large part of the reasoning process for the lawyer.
Lawyers present their arguments, substantiated with citations to precedent cases, to the judge, who again uses precedent in their argumentation.
The judge will cite past cases, much like lawyers do, as part of their argument towards the \textbf{outcome} of a case.
As a judge arrives at the decision, the case at hand will set a binding precedent and the process repeats.\looseness-1

\paragraph{Conclusion.}
Our position is that the best way to build explainable legal outcome prediction models is through the direct use of precedent.
This way, both lawyers and judges can directly interpret and analyze the rationale provided by the model.
Moreover, both lawyers and judges are already trained in reasoning about new cases with respect to old cases, so no additional training would be required.\looseness=-1

\section{Under the Influence}\label{sec:method}
We now turn to the explainability method developed in this paper, which will allow us to identify the exact role precedent plays in the decisions made by neural legal outcome prediction models.\looseness=-1

\newtheorem{example}{Example}

\subsection{Preliminaries}
Legal outcome prediction is a multilabel classification task.
We start with the necessary notation and a generic description of a probabilistic legal outcome prediction model.
Let $\calO = \{\plus, \minus, \outcomenull\}$ be a set of \defn{legal outcomes}, 
    whose elements correspond to a positive, negative and null outcome of a case, respectively.
    The null outcome occurs when a law has not been claimed by either party to the case.
    Let $\mO$ be an outcome-valued random variable with instances $\outcome \in \calO$.
    Let $\calO^K$ be the $K$-times Cartesian product of $\calO$ where $K$ is the number of legal Articles we consider.
    The elements of $\calO^K$ are denoted as $\bo \in \calO^K$.
    Let $\bO$ is a random variable ranging over $\calO^K$.
    We write $\mO_k$ for the random variable, ranging over $\calO$, that corresponds to just the $k^{\text{th}}$ article.
    For instance, if a judge finds that Article 3 has been breached, we write $\mO_k = \plus$.
 Now, let $\vocab$ be an alphabet.\footnote{In principle, the alphabet $\vocab$ can be constructed over either characters, words or sub-word tokens, the last of which is what we do in our experiments.}
    The random variable $\F$ ranges over textual descriptions of \defn{facts}, i.e., $\vocab^*$ for a vocabulary $\vocab$. 
    Elements of $\vocab^*$ are denoted as $\f$.
    For a case concerned with $\artFive$ we might expect the following $\f$.\looseness=-1
    \begin{example}[\href{https://hudoc.echr.coe.int/?i=001-228836}{Article 5 - Right to a fair trial}]
    \noindent Neither the applicants nor any of the other migrants in their group were provided with any information regarding the reason for or the length of their detention. They did not have a chance to appoint a lawyer and were not provided with an interpreter at any point during their detention.
    \end{example}  
We focus on an \defn{article-factored} probabilistic legal outcome predictor, which can be described by the following probability model
\begin{equation}
\begin{aligned}
    \likely(\bO = \bo &\mid \F = \f)  \\
    &= \prod_{k=1}^K \likely(\bigok = \ok \mid \F = \f) \label{eq:probability}
\end{aligned}
\end{equation}
parameterized by $\btheta \in \parameters$.
We further assume that $\parameters$ is a compact subset of $\Rd$.
The training dataset $\dtrain = \{\z^{(n)}\}_{n=1}^N$ is a set of $N$ outcome--fact pairs where $\z^{(n)} = (\f^{(n)}, \bo^{(n)})$.
The test dataset $\dtest = \{\z^{(m)}\}_{m=1}^M$ is a set of $M$ outcome--fact pairs where $\z^{(m)} = (\f^{(m)}, \bo^{(m)})$.
Each training and test pair encodes a relationship between facts and law.
The instance-level negative log-likelihood, i.e., the negative log-likelihood of a \emph{single} training point $\z$, is denoted as $\Loss_{\z}(\btheta) = -\log \likely (\z)$. 
Then, the average cross-entropy for the training set is given by
 \begin{equation}\label{eq:risk}
\Risk(\btheta) = \frac{1}{N} \sum^{N}_{n=1} \Loss_{\z^{(n)}}(\btheta) 
 \end{equation}
A good value for the parameters $\btheta$
may be found by solving the following optimization problem
 \begin{equation}\label{eq:optimization-problem}
     \stheta = \argmin_{\btheta \in \parameters} \Risk(\btheta) + \strength \regularizer(\btheta)
 \end{equation}
where $\regularizer(\cdot)$ is a regularization function with strength $\strength \geq 0$.
Because $\parameters$ is assumed to be compact and $\Risk(\btheta) + \strength \regularizer(\btheta)$ is a continuous function of $\btheta$ the $\argmin$ given in \Cref{eq:optimization-problem} is well-defined.\looseness=-1

\subsection{Influence Functions }
In order to evaluate if and to what extent a model relies on any of the above categories of precedent, we need a way to identify how precedent cases affect the model's decision.
In a legal outcome prediction model, the only cases that can be considered precedent are the training cases. 
A simple technique to measure the effect of a training case $\z$ on a test case $\ztest$ would be to remove $\z$ from the training set $\dtrain$, retrain the model, and take $\z$'s importance to be the change in the loss $\Loss_{\z}(\btheta)$ for a test point $\z$.
However, with over $N = 8000$ training cases, we would have to retrain the model $N$ times to be able to estimate the importance of each case in the training set for the test cases $\ztest$.
This would be prohibitively slow.\looseness=-1

As a tractable alternative to the above, \citet{influence} introduce a simple procedure that allows us to efficiently estimate the effect of a single $\z$ on the model parameters \emph{without} retraining.
We now describe their method, based on influence functions, in detail.
Consider the following optimization problem, a modified version of \Cref{eq:optimization-problem}
\begin{align}
   \params = \argmin_{\btheta \in \bigTheta} \underbrace{\Risk(\btheta) + \strength \regularizer(\btheta)}_{\defeq L_{\regularizer}(\btheta)} + \varepsilon \Loss_{\z}(\btheta)
    \label{eq:newpara}
\end{align}
It is not hard to see that \Cref{eq:newpara} simply modifies the weight associated with $\z$ in the summation found in \Cref{eq:optimization-problem}.
Relying on \citeposs{influence} method, we can approximate the effect of a very small $\varepsilon$ on parameters $\btheta$ by taking a derivative of $\params$ with respect to $\varepsilon$ at evaluating it at $\varepsilon = 0$:
\begin{align}\label{eq:influence}
\!\!\frac{\dev \btheta_z^\star}{\dev \varepsilon}&\Big|_{\varepsilon = 0} \!\!\!\!= \!- \left(\derivative_{\btheta}^2 L_{\regularizer}(\btheta^\star)\right)^{-1} \derivative_{\btheta} \Loss_{\z}(\btheta^\star)
\end{align}
See \cref{app:alsoinfluence} for a derivation of \Cref{eq:influence}.

\paragraph{Influence score.} 
We are not directly interested in the new parameters themselves but rather in the effect of the change in parameters on the loss for a test point $\ztest$. 
We can calculate this influence score using the chain rule:
\begin{align}\label{eq:score}
    \iota(\z, &\ztest) \defeq \frac{\mathrm{d}\Loss_{\ztest}(\btheta_z^\star)}{\mathrm{d}\varepsilon} \Big|_{\varepsilon=0} \nonumber \\ 
                    &= \derivative_{\btheta}\Loss_{\ztest}(\btheta^\star)^\top \frac{\mathrm{d}\btheta_{\z}^\star}{\mathrm{d}\varepsilon} \Big|_{\varepsilon=0}  \quad   \mathcomment{(chain rule)} \\
                    &\overset{(a)}{=}  - \derivative_{\btheta}\Loss_{\ztest}(\btheta^\star)^\top \left(\derivative_{\btheta}^2 L_{\regularizer}(\btheta^\star) \right) ^{-1}\derivative_{\btheta} \Loss_{\z}(\btheta^\star)   \nonumber
\end{align}
where $(a)$ results by plugging in \Cref{eq:influence}.
Thus, we can leverage \citeauthor{influence}'s method to measure the extent to which a precedent case bears on a new legal decision.\looseness-1

\section{Taxonomy of Precedent}\label{sec:corpus}

Precedent is more than just a binary category.
Legal actors use it in a number of different ways to craft their arguments.
To better evaluate how neural models of outcome use precedent, we develop a novel taxonomy of types of precedent.

\subsection{Applied vs. Distinguished}
The simplest way of categorizing the precedential relationship between two cases is to differentiate between application and distinction.\footnote{We focus on the major types of precedent. There are other forms of precedent we do not consider here. Specifically, modification of precedent and overruling. In future work, they would be a natural candidate for further exploration.} 

\paragraph{Applied Precedent.}
We call a precedent case \defn{applied} when a judge applies the precedent from a previous case.
In such instances, the judge has relied on similar reasoning to that of the precedent case.
As a result, the outcome of the case at hand is the same as the outcome of the precedent case.
In practice, these are the precedent cases that a judge has cited as part of their argument, which have the same outcome as the case at hand.\looseness-1

\paragraph{Distinguished Precedent.}
We call a precedent case \defn{distinguished} when a judge distinguishes their decision from a past case.
When a precedent is distinguished, the judge has used the differences between the cases to inform their decision.
In practice, these are the cases that a judge cites but have a different outcome than the case at hand.\looseness-1

\subsection{Positive vs. Negative Precedent}
\citet{valvoda-etal-2023-role} introduce a new perspective on how to model legal outcome prediction. 
By considering both the claims and the outcomes of cases, they define two types of legal outcomes: positive and negative.\footnote{This is not a new concept to lawyers or legal AI theorists. \citet{lawlor} already described these categories of precedent as \emph{pro-precedent} and \emph{con-precedent}.}
Following their logic, we can further sub-divide the applied or distinguished precedent into the two categories below.\looseness-1

\paragraph{Positive Precedent.}
An outcome is positive when a law has been claimed as breached and a judge finds in favor of the claimant.
When a new case cites a positive outcome case, the judge invokes a positive precedent.
In the strictest sense, this means that if a new case arises from exactly the same facts as the precedent case before it, the judge will have to conclude the same law has been equally breached in both cases.
In practice, no two cases are exactly the same. 
Therefore the positive precedent affects all subsequent similar cases by increasing the chances of successfully claiming a breach of the law that has been violated in the precedent case.\looseness-1

\paragraph{Negative Precedent.}
An outcome is negative when a law has been claimed as breached, but the judge finds it has not in fact been broken. 
When a new case cites a negative outcome case, the judge invokes a negative precedent.
Much like with positive precedent, the court is bound by the negative precedent.
Two cases with the same facts must reach the same outcome---positive or negative.
Therefore, the scope of the law contracts through negative precedent.
\looseness-1

\subsection{Scope}

Additionally, there are two axes along which we can define the scope of the precedent we will consider in \cref{sec:results}.\looseness-1

\begin{table}
\centering
\begin{tabular}{lllll}
\toprule
 & \multicolumn{2}{c}{\textbf{Applied}}                                  & \multicolumn{2}{c}{\textbf{Distinguished}}                         \\
 \midrule
                  & \multicolumn{1}{c}{\textbf{Cited}} & \multicolumn{1}{c}{\textbf{Claimed}} & \multicolumn{1}{c}{\textbf{Cited}} & \multicolumn{1}{c}{\textbf{Claimed}}\\
\midrule

 \textbf{Pos.} & 2,012 & 643,377 & 114 & 54,099        \\
\textbf{Neg.} & 296 & 129,239 & 79 & 30,203          \\
   
\bottomrule
\end{tabular}
\caption{Total number of cited cases per category of precedent under the per-case view of the precedent.}
\label{tab:dataset}
\vspace{-15pt}
\end{table}

\paragraph{Claimed vs. Cited.}
There are two ways to uncover the relevant precedent for a case.
The first is to consider all the Articles that the lawyer has \emph{claimed} as violated.
To find these cases, we can look through all past cases and pick out those where the same Articles have been claimed.
Because all cases in our corpus deal with the same subset of law, by the nature of precedent, all past decisions are all binding with respect to future ones.
We call this the \defn{claimed precedent}.
However, in practice, considering all claimed precedent cases yields a large number of precedent cases.
Thus, we also consider a more narrow view.
Specifically, we focus on those cases the judge cites when drafting their judgment.
We call this the \defn{cited precedent}.
Each approach has its limitations.
Under the narrow view (cited precedent), we have a non-exhaustive list of cases that only show us the reasoning of a \emph{single} judge.\footnote{A judgement may be given by a tribunal of judges, each giving a different argument. 
In our dataset, the reasoning towards the decision is reported as a single argument.\looseness=-1}
Under the broader view (claimed precedent), we consider many cases that are not directly relevant to the facts of the case at hand.
The claimed--cited distinction plays a crucial role in our analysis. Correlations are broken down into cited and claimed precedent in \cref{table:results}\looseness-1

\paragraph{Per-case vs. Per-article.}
A lawyer typically claims multiple Articles as violated, and, likewise, a judge may find that multiple Articles were breached.
Thus, when selecting precedent cases, on one hand, we can consider those where all the claimed Articles are shared.
Or, on the other hand, we can consider a piecemeal approach where we select all cases that share \emph{at least} one Article.
We call the former approach the \defn{per-article approach} and the latter the \defn{per-case approach}.
The advantage of the per-case approach is that it selects for highly related cases because the cases have an identical set of claimed Articles.
The advantage of the per-article approach is that it allows for a more detailed analysis---a case might be an important precedent with respect to a judge's decision on only a single Article.
While we report our main results with per-case correlations, we include an analysis of per-article correlations in \cref{sec:article}.\looseness-1

\subsection{Recovering the Taxonomy}
Conveniently, the above taxonomy can be recovered post hoc from the case transcripts.
All we need is to construct a citation network.
We can then identify the relationship between a cited case and the citing case as follows.
First, we identify the outcomes of both the citing case and the cited case.
If the positive outcomes of the cited case are the same as those of the citing case, there is a relationship of \textbf{applied-positive} precedent.
If the negative outcomes are the same, the relationship is that of \textbf{applied-negative} precedent.
To extract cases that constitute a distinguished precedent, we first identify if the case at hand and the cited case share the same claims.
If they do, we move forward to distinguish between positive and negative outcomes.
When the positive outcomes of the cited case are the same as the negative outcomes of the citing case, we label the relationship as \textbf{distinguished-positive}.
Conversely, if the negative outcomes of the cited case are the same as the positive outcomes of the citing case, we label the relationship as \textbf{distinguished-negative}.\looseness=-1

Note that because we consider positive and negative precedent separately, the relationship between any two cases can simultaneously be both positive and negative, though it can not be both applied and distinguished.
Some cited cases might not fall under any of the above categories.
We include an example of how we process cases into the taxonomy in \cref{app:example}.\looseness-1

\section{An Improved ECtHR Corpus}
We modify the ECtHR corpus, a popular legal NLP dataset, for our experiments.
Specifically, we use the scrape of the corpus from \citet{valvoda-etal-2021-precedent}.
An ECtHR case can be separated into two sections: facts and arguments.
The facts of the case, denoted $\f \in \Sigma^*$, contain text describing the events that had brought the claimant to court. 
This section outlines the events that might give rise to a particular breach of law. 
Arguments of the case on the other hand contain a transcription of the reasoning the judge(s) employed in reaching the conclusion over whether or not a breach of law has actually occurred.
The argument section is where the judges cite previous cases, i.e., the precedent.\looseness-1

We first extract the necessary citations to previous cases.
Specifically, for every case in our dataset, we extract the names of all the cases cited in the arguments section.
We use regular expressions for this purpose because citations in the ECtHR corpus follow a regular form.\footnote{ECtHR has issued a note on the rules on case citations \href{https://www.echr.coe.int/documents/d/echr/note_citation_eng}{here}. 
In the ECtHR corpus, the earliest cases are from the year 2001 and the relevant citation styles they follow are the two used between 1999-2015 and from 2016 onwards. 
Both follow a regular format, allowing us to generate our citation dataset.\looseness=-1}
The first author manually inspected the extracted citations to ensure a high-quality citation network.
Then, we remove all the cases in the training set of our dataset that have \emph{not} been cited.
This step ensures that every training case is a precedent of at least one test case.
Then we remove all the cases in the test set of our corpus which do not cite any precedent.
We do this to ensure that, for each case in our test set of the corpus, there is a precedent cited from our training corpus.
Details about the number of cases per Article and the split between the training, validation, and test set are contained in \cref{app:dataset}.
\looseness=-1

\section{Neural Models of Legal Outcome}\label{sec:neural}

We consider two different instantiations of \Cref{eq:probability}, each of which corresponds to a different legal outcome prediction approach from the literature.

\paragraph{\simple.}
Under \citeposs{chalkidis-etal-2022-lexglue} model, legal outcome prediction is treated as a binary classification task with the outcome of a case simplistically assumed to be either $\plus$ or an amalgamation of $\minus$ and $\outcomenull$.
Using a pre-trained language model as an encoder $\enc$, we define the following model:\looseness=-1
\begin{subequations}
\begin{align}
    \bh &= \enc(\f)  \\
    \logits &= \W^{(1)}_k\, \ReLU(\W^{(2)}\,\bh)\\
    \probability(\bigok = \plus \mid \f) &= \sigmoid(\logitsk)_{+} 
\end{align}
\end{subequations}
where the multi-layer perceptron (MLP) is parametrized by $\W^{(1)}_k \in \R^{1 \times d_2}$ and $\W^{(2)} \in \R^{d_2 \times d_1}$.
The input $\bh \in \R^{d_1}$ is a high-dimensional representation of the facts computed by the encoder.\looseness=-1

\begin{table*}
\centering
\begin{tabular}{llllrrrrr}
\toprule
&                    & &                   & \multicolumn{2}{c}{\textbf{Applied}}                                  & \multicolumn{2}{c}{\textbf{Distinguished}} &  {\textbf{Overall}}                          \\
\midrule
\textbf{Type} & \textbf{Language Model}                    & $\mathbf{F_1}$ &                   & \multicolumn{1}{c}{\textbf{Cited}} & \multicolumn{1}{c}{\textbf{Claimed}} & \multicolumn{1}{c}{\textbf{Cited}} & \multicolumn{1}{c}{\textbf{Claimed}} &  \\
\midrule

Simple & BERT  & 0.64 & \textbf{Positive} & 0.013 & 0.148 & -0.002 & -0.012  &   0.137          \\
                                    & & & \textbf{Negative} & 0.004 & 0.037 & -0.002 & -0.023          \\
     & LEGAL-BERT & 0.67 & \textbf{Positive} & 0.013 & 0.180 & -0.003 & -0.026   & 0.157                \\
                                         & & & \textbf{Negative} & 0.003 & 0.027 & -0.002 & -0.023                     \\       
Joint & BERT & 0.66 & \textbf{Positive} & 0.008 & 0.079 & 0.001 & 0.020 & 0.084         \\
                                        & & & \textbf{Negative} & 0.003 & 0.023 & 0.001 & 0.002                  \\
     & LEGAL-BERT & 0.68 & \textbf{Positive} & 0.008 & 0.078 & 0.000 & 0.020 & 0.085\\
                                                 & & & \textbf{Negative} & 0.004 & 0.031 & 0.000 & 0.004  \\       
   
\bottomrule
\end{tabular}
\caption{We report per-case Spearman's $\rho$ between influence scores and our taxonomy of precedent, $F_{1}$ scores for the models under consideration and the \emph{overall} correlation between influence scores and any type of precedent.}
\label{table:results}
\end{table*}

\paragraph{\joint.}
We also implement \citeposs{valvoda-etal-2023-role} model, which treats outcome prediction as a three-way classification task:
\begin{subequations}
\begin{align} 
    \label{eq:joint_model}
    \bh &= \enc(\f)  \\ 
    \logits &= \W^{(3)}_k\, \ReLU(\W^{(4)}\,\bh) \\
    \probability(\bigok = \ok \mid \f) &= \softmax(\logitsk)_{\ok} 
\end{align}
\end{subequations}
where $\W^{(3)}_k \in \R^{3 \times d_2}$ and $\W^{(4)} \in \R^{d_2 \times d_1}$ are the parameters of the MLP.
The advantage of this approach is that it directly models both positive and negative outcomes, which are closely related to positive and negative precedent.
Thus, we expect the {\joint} to be better at utilizing them.

\paragraph{Choice of $\enc$.}
For both models, we consider two choices for the encoding $\enc : \Sigma^* \rightarrow \mathbb{R}^{d_1}$: (1) BERT \citep{devlin-etal-2019-bert} and (2) LEGAL-BERT \citep{chalkidis-etal-2020-legal}.
The full experimental details, e.g., hyperparameters, are given in \cref{app:details}.

\definecolor{green}{RGB}{84, 168, 104}
\definecolor{orange}{RGB}{221, 132, 82}
\definecolor{blue}{RGB}{76, 114, 177}

\section{Results}\label{sec:results}
We report our results in terms of Spearman's $\rho$.
Overall, we find a weak positive correlation between judges' precedent and influence scores---the best correlation we find is $0.180$.
However, due to the fine-grained taxonomy, the results vary between different types of precedent.
Inspecting the main results, contained in \cref{table:results}, we 
make the following six findings.\looseness-1

\paragraph{Finding \#1.}
We observe that influence scores correlate more strongly with applied precedent than distinguished precedent.
In fact, the distinguished precedent is negatively correlated with the influence scores in the case of the \simple.
Under the best \joint, the highest correlation we find between the distinguished precedent and influence scores is nearly $4\times$ lower ($0.020$) when compared to the highest correlation we obtain for the applied precedent ($0.079$).
Therefore, we conclude that current legal outcome prediction models struggle with distinguished precedent.\looseness-1

\paragraph{Finding \#2.}
We find negative precedent correlates less strongly with influence scores than positive precedent.
In the case of applied precedent, we observe a correlation with positive precedent of $0.180$, which is more than $6\times$ larger than the correlation we observe with negative precedent ($0.027$).
Therefore, we conclude the models struggle to utilize negative precedent in comparison.\looseness-1

\paragraph{Finding \#3.}
When analyzed through the lens of influence scores, neither model correctly emulates the full spectrum of precedential reasoning.
While the influence scores under the \joint{} correlate positively with every type of precedent, we observe an overall correlation of $0.085$, which is weaker than the overall correlation of $0.157$ that we observe with the \simple{}'s influence scores.
Because the \simple{} is negatively correlated with distinguished precedent, neither of the models under consideration seems to be able to robustly reason with the full set of precedential operations.\looseness-1

\paragraph{Finding \#4.}
We observe that pre-training on legal data is beneficial for improving precedent correlations.
The LEGAL-BERT-based models exhibit higher correlations with precedent when compared to the BERT-based models.
Under the \simple, in particular, we see an increase from a correlation of $0.137$ to a correlation of $0.157$ when we use LEGAL-BERT.\looseness-1

\paragraph{Finding \#5.}
When we compare the correlation for cases a judge has cited with the correlation for claimed precedent cases, we observe that the latter approach magnifies the results. 
In the case of a positive correlation, the correlations become more positive. 
In the case of a negative correlation, the correlations become more negative. 
The effect is considerable---in many instances, the difference is more than an order of magnitude.\looseness-1

\paragraph{Finding \#6.}
We find that higher $F_1$ scores do not directly translate to more human-like legal reasoning.
Where the \joint s outperform the \simple s in terms of $F_1$, the overall $\rho$ is higher for the \simple s.
Overall, the worst model in terms of $F_1$ (Simple BERT: $0.64$ $F_1$) exhibits a correlation of $0.14$ with precedent. 
Whereas the best model in terms of $F_1$ (Joint LEGAL-BERT: $0.68$ $F_1$) exhibits a correlation of merely $0.085$ with the precedent.\looseness-1

\paragraph{Conclusion.}
In summary, we find that current models do well at encoding one particular type of precedent, applied-positive precedent, and not so well at the other types of precedent.
We believe the culprit is the lack of sophisticated legal reasoning.
To distinguish a case is to identify how two seemingly similar cases differ.
This requires in-depth legal knowledge.
Our work indicates that this level of legal understanding is beyond the capabilities of the current neural models of legal outcome prediction models.\looseness=-1

\begin{figure*}
    \centering
    \subfloat{{\includegraphics[width=7.5cm]{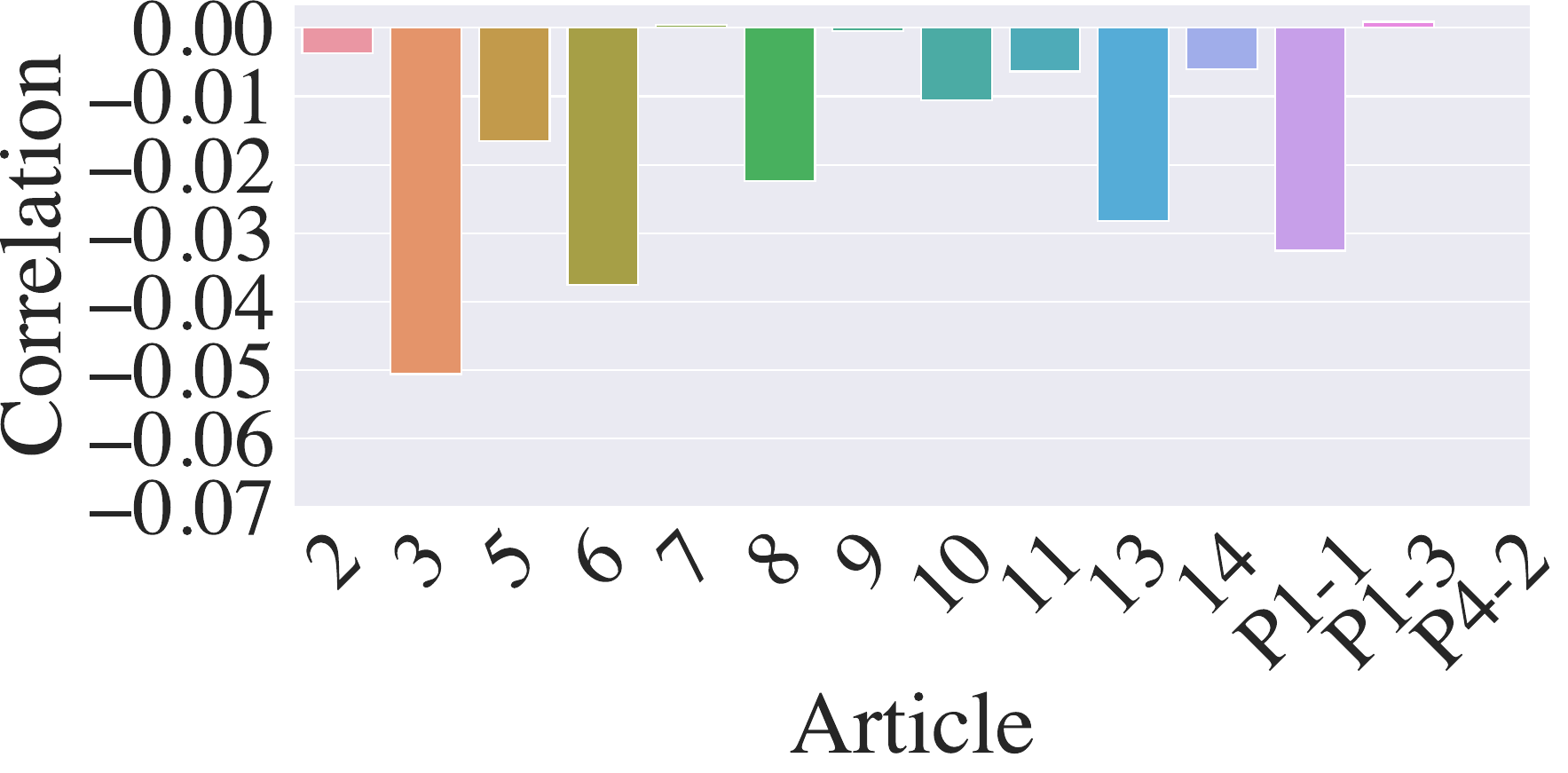} }}%
    \qquad
    \subfloat{{\includegraphics[width=7.5cm]{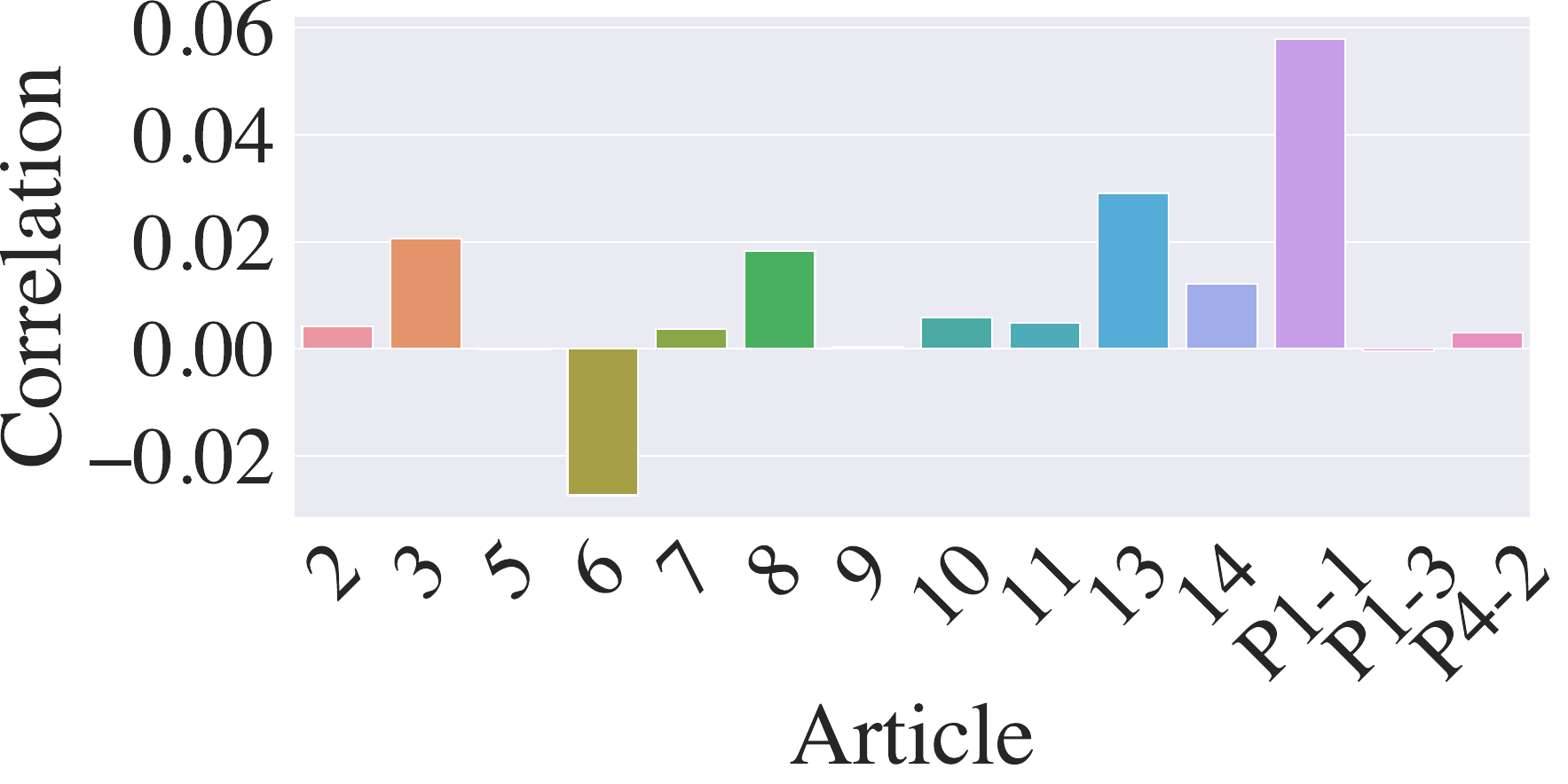} }}%
    \caption{Per-article Spearman correlation between influence functions and cited negative-distinguished precedent.
    The \simple{} is on the left and the \joint{} is on the right; both models are LEGAL-BERT based.}
    \label{fig:article}%
\end{figure*}

\section{Further Analysis}\label{sec:discussion}
We extend our analysis with the following three studies to better understand the models' behavior.
First, we analyze the precedent from the perspective of the model. 
Then, we analyze the model behavior per individual legal Article. 
Finally, in \cref{app:metric}, we develop a novel metric, which confirms the trends we report in \cref{sec:results} from the perspective of precedent retrieval.\looseness-1

\subsection{Model-based Precedent}\label{sec:alternativeview}
Current legal outcome predictors often fail to predict the correct outcomes.
Therefore, the weak correlations reported above might simply be due to the model relying on the precedent which supports the outcome the model predicts, rather than the outcome a human judge has reached.
To investigate this possibility, we develop two ways of selecting precedent in line with the models' prediction of the outcome.
First, we compute correlations for only those instances where the model has \textbf{correctly predicted} the outcome.
Second, we select the precedent cases by taking the model outcome prediction as the ground truth. 
We call this approach \textbf{model-based}.
Under both approaches, we study the same types of correlations as in \cref{sec:results}.
The overall Spearman correlations are reported in \cref{table:alternative}.\looseness-1

We find that the correlation with precedent is higher for correctly predicted cases.
The best overall correlation for these instances improves from $0.157$ to $0.183$, a $17\%$ improvement. 
On the other hand, the model-based results reveal a decrease in correlation with precedent. 
The best model under this paradigm achieves a correlation of only $0.011$, more than an order of magnitude worse than the best model in \cref{table:results}. 
This suggests that the models agree with a human judge's choice of precedent more than with the precedent that would be consistent with the model's own predictions. 
This finding casts further doubt on the alignment of the models' reasoning with the doctrine of precedent.
We report full results in \cref{app:modelbased}.\looseness=-1

\begin{table}[]
\begin{tabular}{lllll}
\toprule
\textbf{Type}  & \textbf{LM} & \textbf{Correct} & \textbf{Model} \\
\midrule
Simple & BERT       & 0.149  & 0.007 \\
       & LEGAL-BERT & 0.183  & 0.032 \\
Joint  & BERT       & 0.138  & 0.010  \\
       & LEGAL-BERT & 0.148  & 0.011  \\
\bottomrule
\end{tabular}
\caption{Spearman's correlations for the correctly predicted cases and model-based precedent. 
The correlations are based on \emph{overall} types of precedent.}
\label{table:alternative}
\end{table}

\subsection{Per-article Analysis}\label{sec:article} 
So far, we have considered only per-case correlations. We now turn to the per-article results.
We report our full per-article results in \cref{app:article}.
In general, we observe the same trends as those set out in \cref{sec:results}. 
Nonetheless, we find interesting results for the negative distinguished precedent; see \cref{fig:article}.
We discuss these results in more detail below.\looseness-1

Under the \simple, we observe that those Articles that are more heavily represented in the training data exhibit a lower correlation with negative distinguished precedent in comparison with those that are less heavily represented.
For example, Articles $3$, $6$, $8$, $10$, $13$ and P1-1, each of which has a large number of examples in the training set, correlate more negatively with negative distinguished precedent than their less well-represented counterparts.
Therefore, we contend that it is unlikely that introducing additional training data will solve the problem of limited precedential reasoning in neural models of legal outcome.
In fact, scaling up the data seems to result in less human-like behavior.
Instead, our results suggest that we need to develop new models with a better legal inductive bias to address this issue.\looseness-1

When it comes to distinguished precedent, {\joint} exhibits some aspects of such inductive bias.
For most Articles, and Articles $8$ and $13$ in particular, the model seems to utilize distinguished precedent. 
Nonetheless, for Article $6$, which is the most represented Article in the dataset, we still observe a negative correlation.
While {\joint} appears to be a step in the right direction in some respects, it is still far from using the precedent in a way a human judge would.\looseness-1

\section{Related Work}\label{sec:related}
In the wider NLP context, there is a considerable amount of research aimed at developing methods for unraveling what LMs know about language \cite{alain2016understanding, shi-etal-2016-string,ettinger-etal-2016-probing,bisazza-tump-2018-lazy,liu-etal-2019-linguistic, pimentel-etal-2020-information}.
Sub-fields of NLP dealing with sensitive information, such as biomedical NLP \cite{garcia-olano-etal-2021-biomedical, interpretablebio1, interpretablebio2, sarker, mullenbach-etal-2018-explainable}, have set out to build their own explainable AI models to address the concerns about applying a black-box model to sensitive data.\looseness-1

In the past, legal AI researchers working on symbolic methods built explainable models of legal reasoning \cite{hypo, aleven1997teaching, rissland1991cabaret, branting1991}.
However, the most recent crop of models utilizing modern NLP methods are inherently opaque.
Regardless of whether the domain is Chinese \cite{feng-etal-2022-legal}, Japanese \cite{yamada2023japanese}, Indian \cite{malik-etal-2021-ildc} or UK law \cite{Valvoda18}, the machine learning models underlying the work in the area are not easily interpretable.

To address this shortcoming, straightforwardly adopting existing methodology from wider NLP is insufficient.
After all, how well a model knows syntax can only go so far in explaining its legal decisions.
Using more general methods, such as attention maps, is not particularly viable either.
For instance, using attention to find where the model \emph{looks} at legal facts does not provide a legally salient answer as to what the decision is based on \cite{mullenbach-etal-2018-explainable}.

Setting aside the question of whether attention can be used for explainability in the first place \cite{serrano-smith-2019-attention, bastings-filippova-2020-elephant, meister-etal-2021-sparse, bibal-etal-2022-attention}, models of legal outcome prediction are trained to make decisions based on the facts of a case alone. 
While the facts of the case are important, inspecting the attention distribution over them can never provide a sufficient explanation. 
This is because lawyers and judges do not argue by highlighting which facts played a role in their decision.
Instead, in their arguments, they justify their decision by relating the facts at hand to previous case decisions.
This makes our work different from prior research in two aspects.
(1) Where the existing Legal AI work focuses on providing explanations by identifying relevant tokens \cite{xu-etal-2023-dissonance}, sentences \cite{malik-etal-2021-ildc} or paragraphs \cite{chalkidis-etal-2021-paragraph} of the facts of the case, we find the entire relevant cases.
(2) Instead of looking inside the input to the model, i.e., at the facts, we look at the training data of the model.\looseness-1

\section{Conclusion}\label{sec:conclusion}
We introduce a method based on influence functions for legal explainability.
We created a taxonomy of precedent, used it to annotate the ECtHR dataset and ran a series of experiments to identify what precedential operations might be learned, or not, by outcome prediction models.
We find that the models we train rely on one type of precedent in particular: applied precedent.
This makes them very unlike human judges, who employ a full spectrum of precedential reasoning to justify the outcome of a case.
To construct useful models of legal outcomes, we need to develop legally faithful models of law.
We hope that our methodology can help in the search for such models.\looseness-1

Indeed, we see two concrete avenues toward legal NLP models that could better align with human precedential reasoning.
(1) We can focus on creating more representative datasets of the tasks we would like to model.
The facts in current datasets contain superficial indicators of the outcome---they are written as post-hoc justification.
Using datasets built around legal briefs, instead of cases, could be a step towards mitigating this issue \cite{valvoda2023ethics}.
(2) We can better incorporate the court processes in our neural models.
To induce precedential reasoning, a natural step would be to incorporate precedent retrieval as part of the model architecture.
Much like a judge, the model should be able to explicitly select the precedent that its decision is based on.\looseness-1

\section*{Limitations}
There are a number of limitations to our work which we would like to acknowledge.
First, our work is limited to a study of the English version of the ECtHR corpus.
Future work should investigate the French translations of the caselaw, which is available for all cases in the corpus.
Additionally, one could experiment with the translations available in other languages, since every case is also translated into the official language of the defendant.
Second, there are potential limitations to using our methodology in different jurisdictions.
We report the details of our dataset and experiments in \cref{app:dataset} and \cref{app:details}, respectively.
Training of the models requires a sufficiently large dataset which could be a limiting factor for smaller jurisdictions with fewer training cases available.

\section*{Ethics Statement}\label{sec:ethics}

The models and methods in this paper are intended purely for scientific research.
While we believe that one day legal AI could be a valuable tool in speeding up the legal process, the current state of the research, including ours, is far from being ready for real-world deployment.
We would therefore caution anyone from using our work towards automating away the job of a human judge or lawyer.
Our results in fact suggest that the current models behave unlike human judges when it comes to precedent. 
This makes them highly unsuitable for real-world deployment.
See our work on legal ethics for a more detailed critique of the current state of Legal AI \cite{valvoda2023ethics}.\looseness-1

\section*{Aknowledgments}
This research is funded by the Nordic Programme for Interdisciplinary Research Grant 105178 and the Danish National Research Foundation Grant no. DNRF169. We would also like to thank Professor Ken Satoh for the fruitful discussions and feedback.

\bibliography{anthology.bib, custom.bib}
\bibliographystyle{acl_natbib}

\appendix

\section{Dataset Details}\label{app:dataset}
We report the breakdown of the dataset in \cref{table:corpus} below.
As you can see, the dataset is highly imbalanced, which is why we conduct our per Article analysis in \cref{sec:discussion}.

\begin{table}[th]
\centering
\begin{tabular}{lccc}
\toprule
      & \textbf{Train} & \textbf{Validation} & \textbf{Test} \\
\midrule
\textbf{No. of Cases} & 8344 & 904 & 913 \\
\midrule
\textbf{Art 2} & 463 & 56 & 55 \\
\textbf{Art 3} & 1223 & 169 & 173 \\
\textbf{Art 5} & 1255 & 176 & 154 \\
\textbf{Art 6} & 4434 & 276 & 272 \\
\textbf{Art 7} & 24 & 2 & 3 \\
\textbf{Art 8} & 638 & 76 & 111 \\
\textbf{Art 9} & 37 & 4 & 4 \\
\textbf{Art 10} & 268 & 39 & 66 \\
\textbf{Art 11} & 102 & 29 & 33 \\
\textbf{Art 13} & 1124 & 77 & 71 \\
\textbf{Art 14} & 119 & 17 & 11 \\
\textbf{P1-1}  & 1307 & 130 & 116 \\
\textbf{P1-3} & 43 & 4 & 1 \\
\textbf{P4-2} & 31 & 3 & 4 \\
\bottomrule
\end{tabular}
\caption{The number of cases in the precedent corpus and the representation of each ECtHR Article.}
\label{table:corpus}
\end{table}

\section{An Example of Our Taxonomy}\label{app:example}
To better understand how we taxonomize precedent, let us consider an imaginary case $\bo$ with five precedent candidate cases $\bo^{'}_{1}, \ldots,\bo^{'}_{5}$.
For the sake of simplicity, we will set these cases in an imaginary legal system, which only has four laws (i.e., $K = 4$).
Below, we illustrate how we would assign the type of precedent to each candidate case:
\begin{align}\label{eq:comparison}
    \bo &= [\outcomenull,+,-,\outcomenull] \quad \!\!\mathcomment{(Case at hand)} \nonumber \\
    \bo'_{1} &= [\outcomenull,+,\outcomenull,\outcomenull] \quad \mathcomment{(Applied positive)} \nonumber \\
    \bo'_{2} &= [\outcomenull,\outcomenull,-,\outcomenull] \quad \mathcomment{(Applied negative)} \nonumber \\
    \bo'_{3} &= [\outcomenull, \outcomenull, +, \outcomenull] \quad \mathcomment{(Distinguished positive)} \nonumber \\
    \bo'_{4} &= [\outcomenull,-,\outcomenull,\outcomenull] \quad \mathcomment{(Distinguished negative)} \nonumber \\
    \bo'_{5} &= [\outcomenull, \outcomenull, \outcomenull, +]\quad \mathcomment{(No precedent)} \nonumber
\end{align}

\section{A New Evaluation Metric}\label{app:metric}
We now develop a novel evaluation metric to determine the degree to which a case's influence corresponds to whether or not it was used as a precedent by the model.
To map a case's influence to the binary decision of whether or not the case was used as precedent, we design a simple classifier.
And, to deal with the imbalanced nature of the classification (there are far more non-relevant cases than relevant ones), we evaluate the classifier with $F_1$.
We omit this analysis from the main text because the conclusions drawn from it are nearly identical to those made in the main with more standard methods.

We now define the classifier formally.
Given an influence score $\iota(\z, \ztest) = x$, we construct the classifier:
\begin{equation}\label{eq:classifier}
    p(C_n = 1) = \frac{\exp(ax_n + b)}{1 + \exp(ax_n + b)}
\end{equation}
where $C_n = 1$ indicates the $n^{\text{th}}$ training case is considered precedent.
The function to the right of \Cref{eq:classifier} is a sigmoid function.
Thus, we know that the classification boundary induced by \Cref{eq:classifier} is:
\begin{equation}
ax + b \geq 0 \Longrightarrow x + \frac{b}{a} \geq 0 \Longrightarrow x \geq -\frac{b}{a}
\end{equation}
Therefore, under \Cref{eq:classifier}, any parameters $a$ and $b$ give us a real value $-\frac{b}{a}$ such that a case is predicted to be precedent iff its influence is greater than $-\frac{b}{a}$. 
To learn the two parameters $a = \theta_{0}, b = \theta_{1}$, 
we optimize the log-likelihood:
\begin{equation}
\sum_{n=1}^N \log p(C_n = 1) + \tiebreaker ||\btheta||_2^2
\end{equation}
where $\tiebreaker > 0$ is a small positive number used to break ties.\looseness=-1

\paragraph{Precedent Classifier Results.} 
We report the results of the above classifier in \cref{tab:bound}. 
In comparison to the random baseline, the influence scores are consistently a better predictor of whether or not a case has been used as a precedent. 
This finding holds over all models under consideration.
The results also confirm our previous findings---\citeposs{valvoda-etal-2023-role} model underperforms \citeposs{chalkidis-sogaard-2022-improved} model.
Furthermore, we find that a classifier on top of a model's influence scores does not score highly on the precedent prediction task.
To contextualize these results, however, we note the results of Task 1 of the past two years of COLIEE competition, which is similar to our precedent retrieval above (but over a different dataset), indicate that this is a difficult problem to solve even for legal information retrieval models.
For context, the best-performing teams in 2021 and 2022 scored $0.19$ $F_1$ and $0.37$ $F_1$ respectively on the data from the Federal Court of Canada case law \cite{coliee21, coliee22}.

\begin{table}[]
\begin{tabular}{llrrr}
\toprule
       &            & \textbf{Model} & \textbf{Rand.}  & \textbf{Gain}  \\
\midrule
Simple & BERT   & 0.25 & 0.18 & +0.07 \\
       & L-BERT & 0.27 & 0.18 & +0.09\\
Joint  & BERT   & 0.21 & 0.18 & +0.03 \\
       & L-BERT & 0.21 & 0.18 & +0.03 \\
\bottomrule
\end{tabular}
\caption{The $F_1$ scores achieved by our models on the precedent retrieval task on the ECtHR corpus.}
\label{tab:bound}
\end{table}\ryan{Which dataset is this in Table 5? \response{josef} I've reported the dataset. It is not.}

\section{Experimental Details}\label{app:details}

In this section, we provide the details of our experimental setup.

\paragraph{Language Model Encoders.}
We use two Transformer-based \cite{attentionisall} pre-trained language models for our experiments.
Following \citet{chalkidis-etal-2019-neural, chalkidis-etal-2022-lexglue} 
we choose to work with base versions of BERT and LEGAL-BERT architectures.\looseness-1

\begin{itemize}[leftmargin=*]
    \item We use BERT \cite{devlin-etal-2019-bert} because it is both a popular language model for benchmarking NLP models and also places at the top of LexGLUE \cite{chalkidis-etal-2022-lexglue} leaderboard for Task A---outcome classification, which most closely resembles what our models are trained for. It also performs the best in the recent reformulation of the task by \citet{valvoda-etal-2023-role}.

    \item Because BERT is not specifically trained on legal text, we additionally use LEGAL-BERT \cite{chalkidis-etal-2020-legal}, which has been fine-tuned on legal data. This allows us to see if exposure to legal language might affect the use of precedent by the model.
\end{itemize}

\paragraph{Training Procedure.}

The models are trained end-to-end by minimizing the cross-entropy between the model and the empirical distribution of the training data.
We conduct a light hyper-parameter search over the learning rate $\{3e^{-4}, 3e^{-5}, 3e^{-6}\}$, size of hidden states $d_2$ in the MLP $\{50, 100, 200, 300\}$ and dropout rate $\{0.1, 0.2, 0.3, 0.4\}$.
We train our models for a maximum of ten epochs with early stopping on validation loss.\footnote{We stop training when validation loss over an epoch stops improving.}
The models are trained for a maximum of 2 hours on a single GPU.
We report our results for the model that achieves the lowest loss on the validation set.

\paragraph{Influence Implementation.}
We rely on a PyTorch \cite{pytorch} implementation of \citeposs{influence} method,\footnote{\href{https://github.com/nimarb/pytorch_influence_functions}{A GitHub link to the implementation we rely on.}} a re-implementation of the original TensorFlow code by Pang Wei Koh.\footnote{\href{https://github.com/kohpangwei/influence-release}{A GitHub link to the original implementation by Koh.}}\looseness-1

\paragraph{Calculating Correlations.}
We report our results in terms of Spearman's $\rho$.
By calculating the influence score of every case in the training corpus on every decision in the test corpus, we can begin to quantify the use of precedent in neural models of legal outcome prediction by studying the correlation of the influence scores with our taxonomy of precedent; see \cref{sec:corpus}.
We store the scores as a vector $\s \in \R^M$, where $M$ is the number of train cases times the number of test cases.
Then, for each of our four categories of precedent, we create a binary vector $\cs \in \{0, 1\}^M$.
For each element in $\s$, $\cs$ encodes whether the training case is that particular type of precedent $1$ or not $0$.
We compute the correlation between $\cs$ and $\s$ and report Spearman's $\rho$.
Additionally, we compute the overall correlation over all types of precedent.
To do this, we simply assign $1$ in $\s$ to a case if it falls under \emph{any} of the precedential categories.\looseness-1

\section{All Per-Article Results}\label{app:article}

Here we report the results of our per-article analysis.
The correlations for individual legal Articles follow the trends reported in \cref{sec:results}.\looseness=-1

\begin{figure}
    \centering
    \subfloat[Applied-Positive]{{\includegraphics[width=7.5cm]{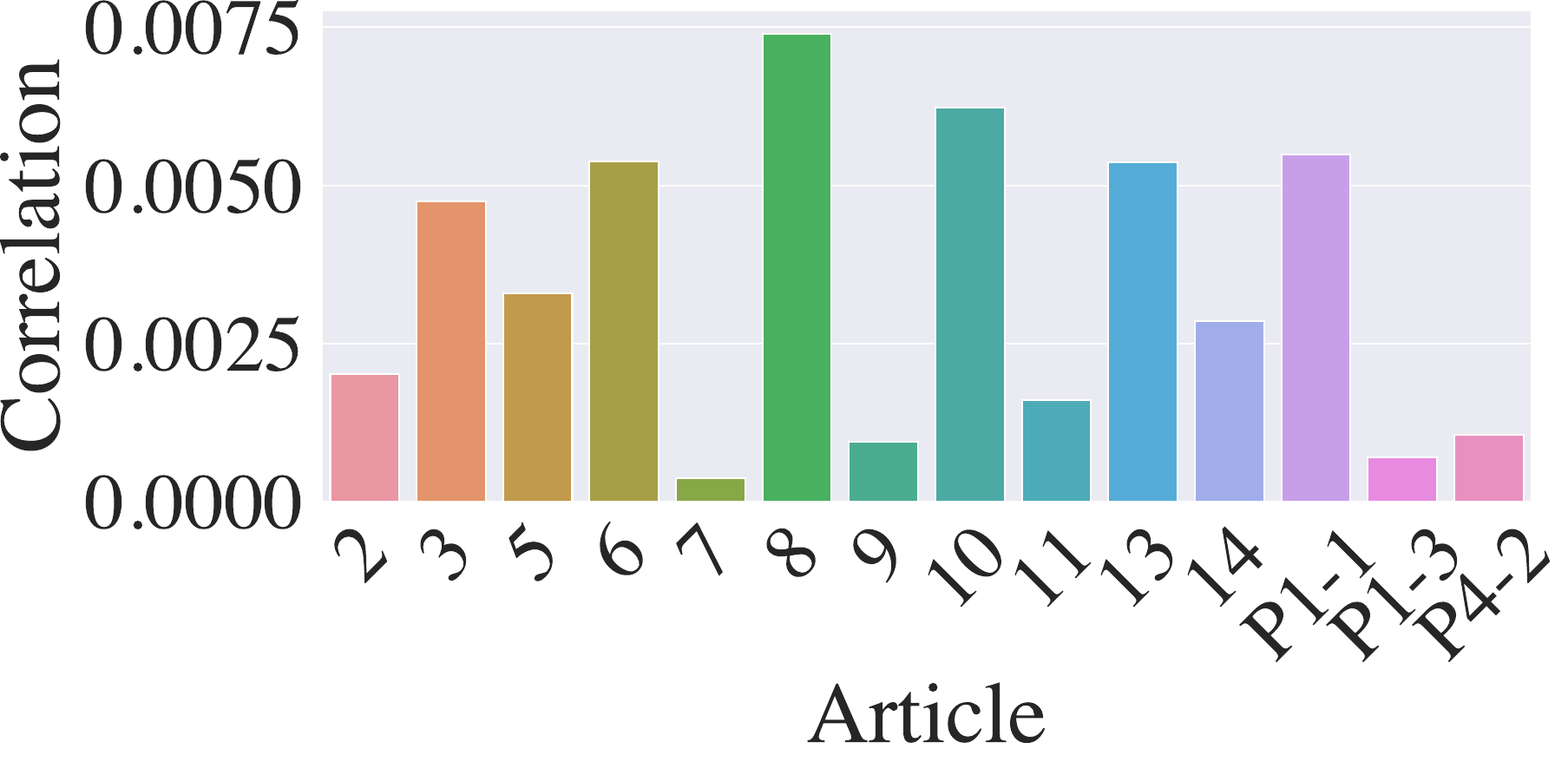} }}\\
    \subfloat[Applied-Negative]{{\includegraphics[width=7.5cm]{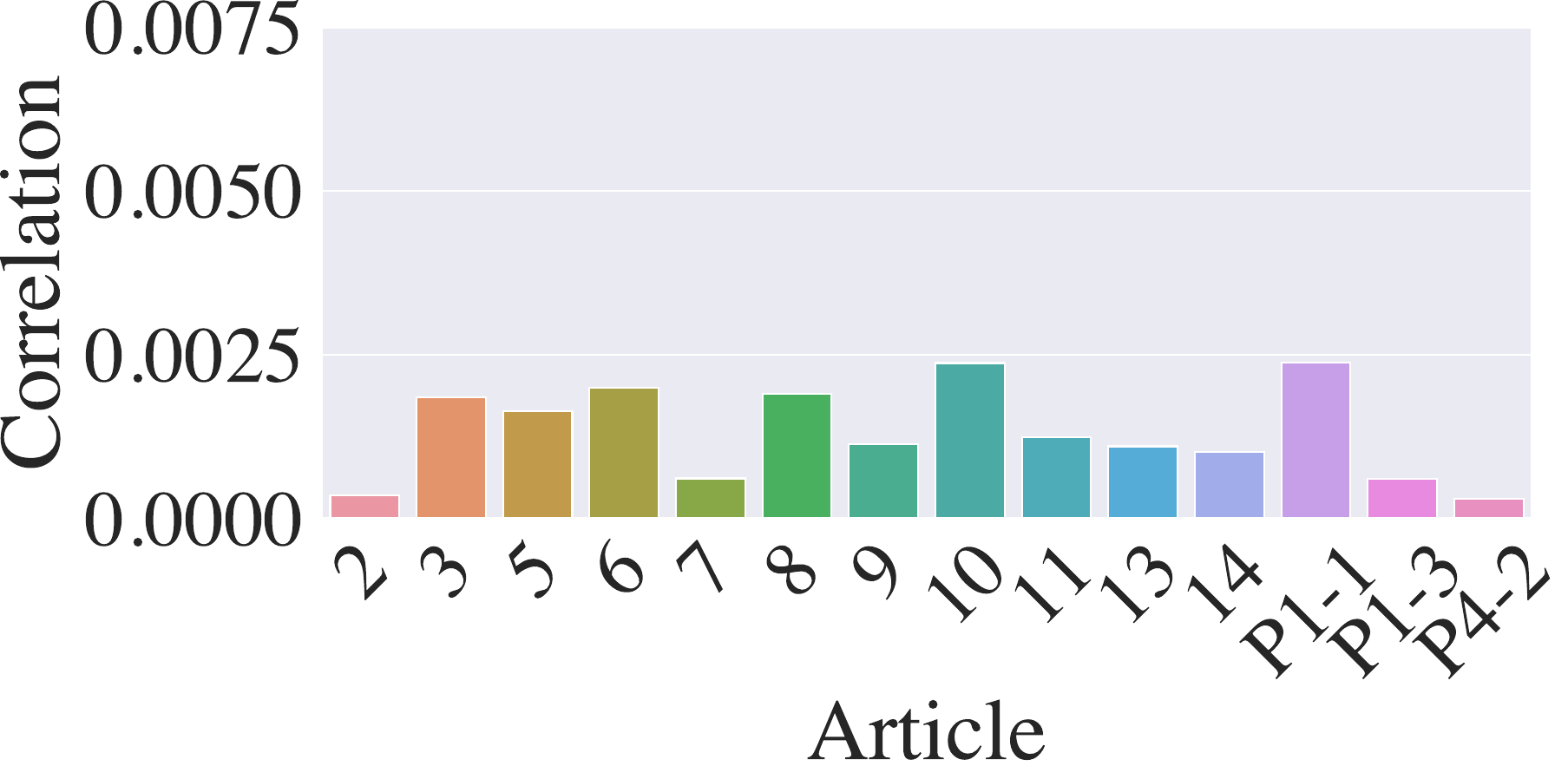} }}\\
    \subfloat[Distinguished-Positive]{{\includegraphics[width=7.5cm]{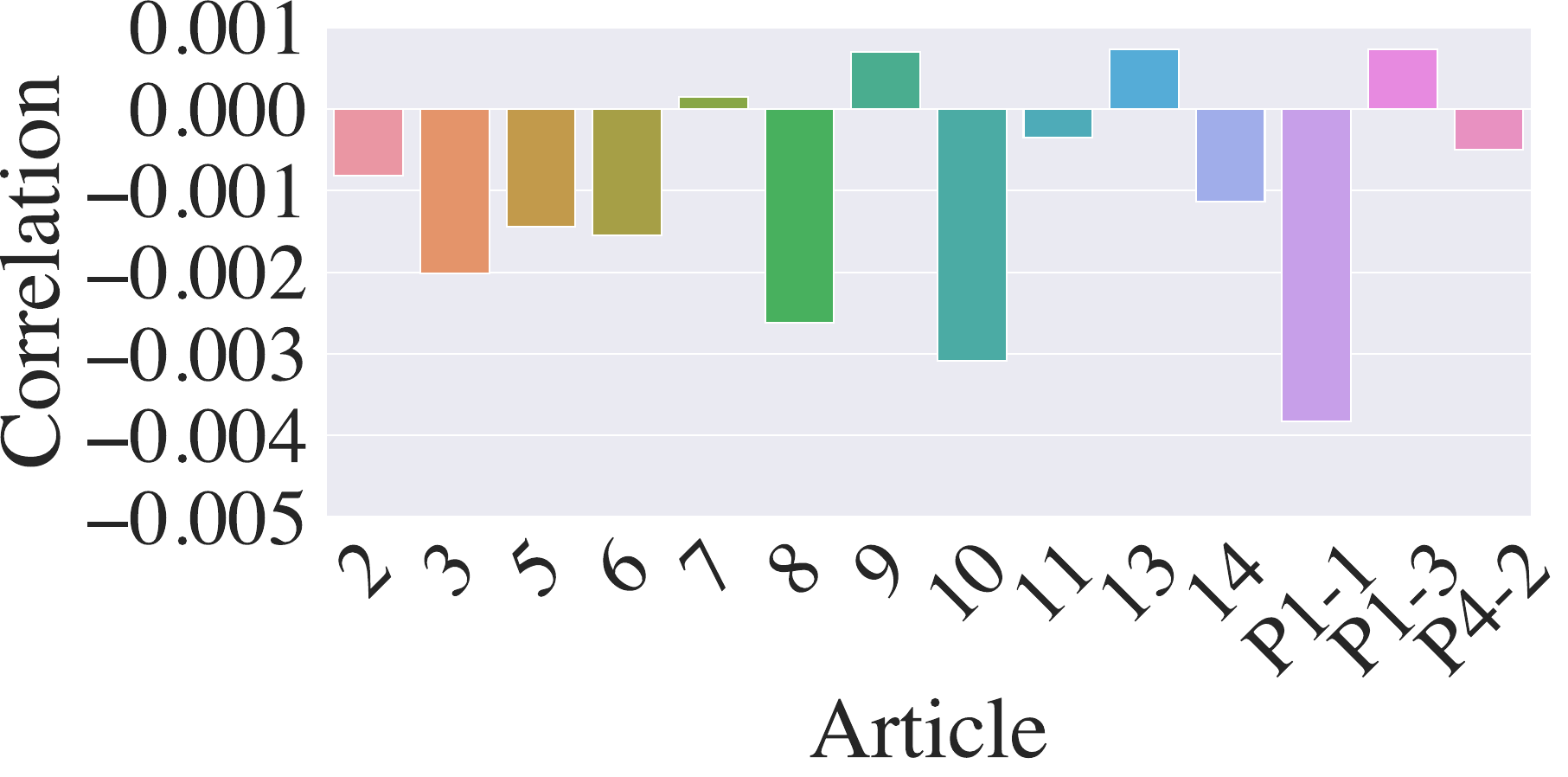} }}\\
    \subfloat[Distinguished-Negative]{{\includegraphics[width=7.5cm]{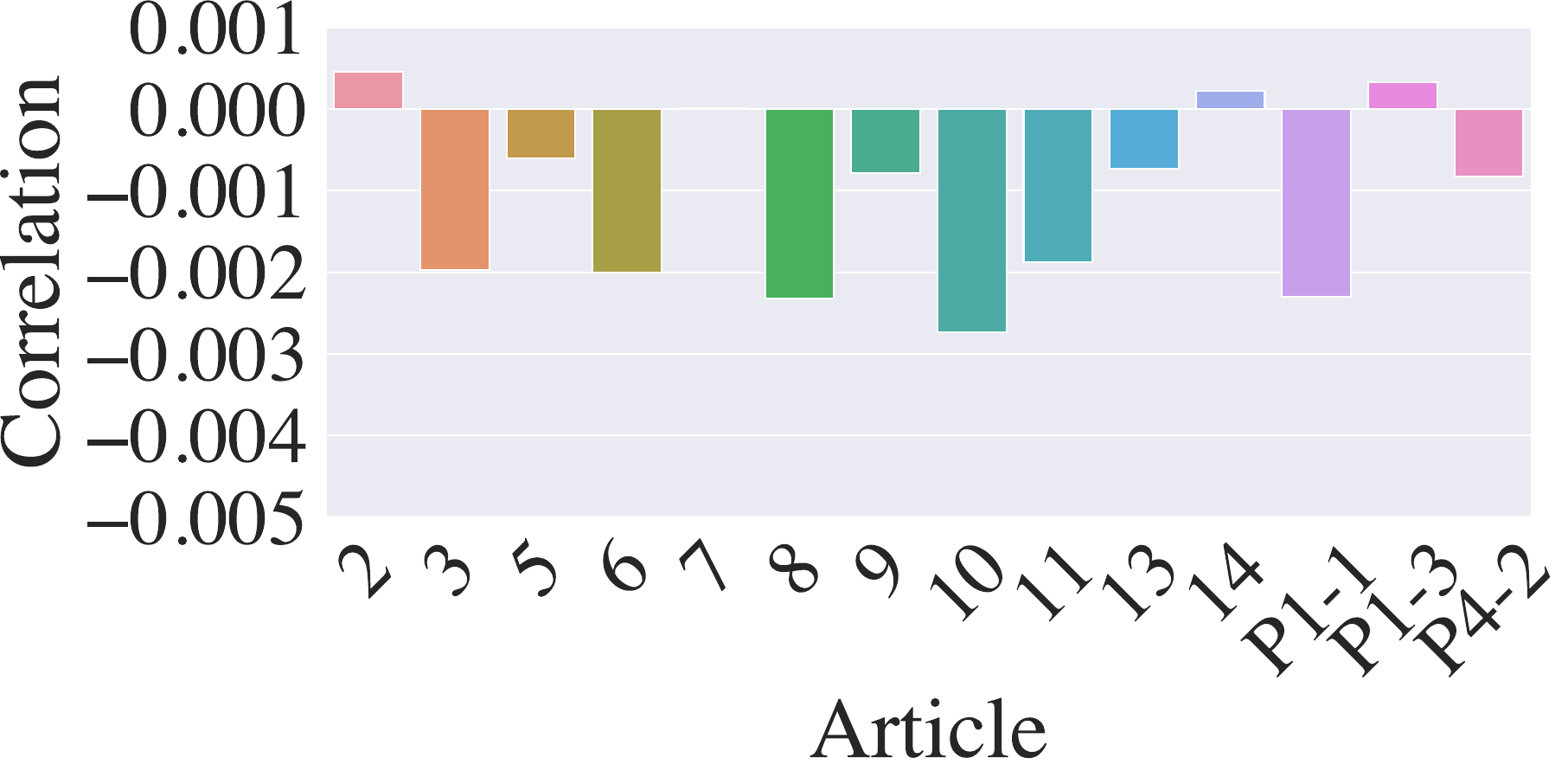} }}\\
    \caption{Spearman's correlation for cited precedent under the \simple{} with a BERT encoder..}
    \label{fig:article}%
\end{figure}

\begin{figure}
    \centering
    \subfloat[Applied-Positive]{{\includegraphics[width=7.5cm]{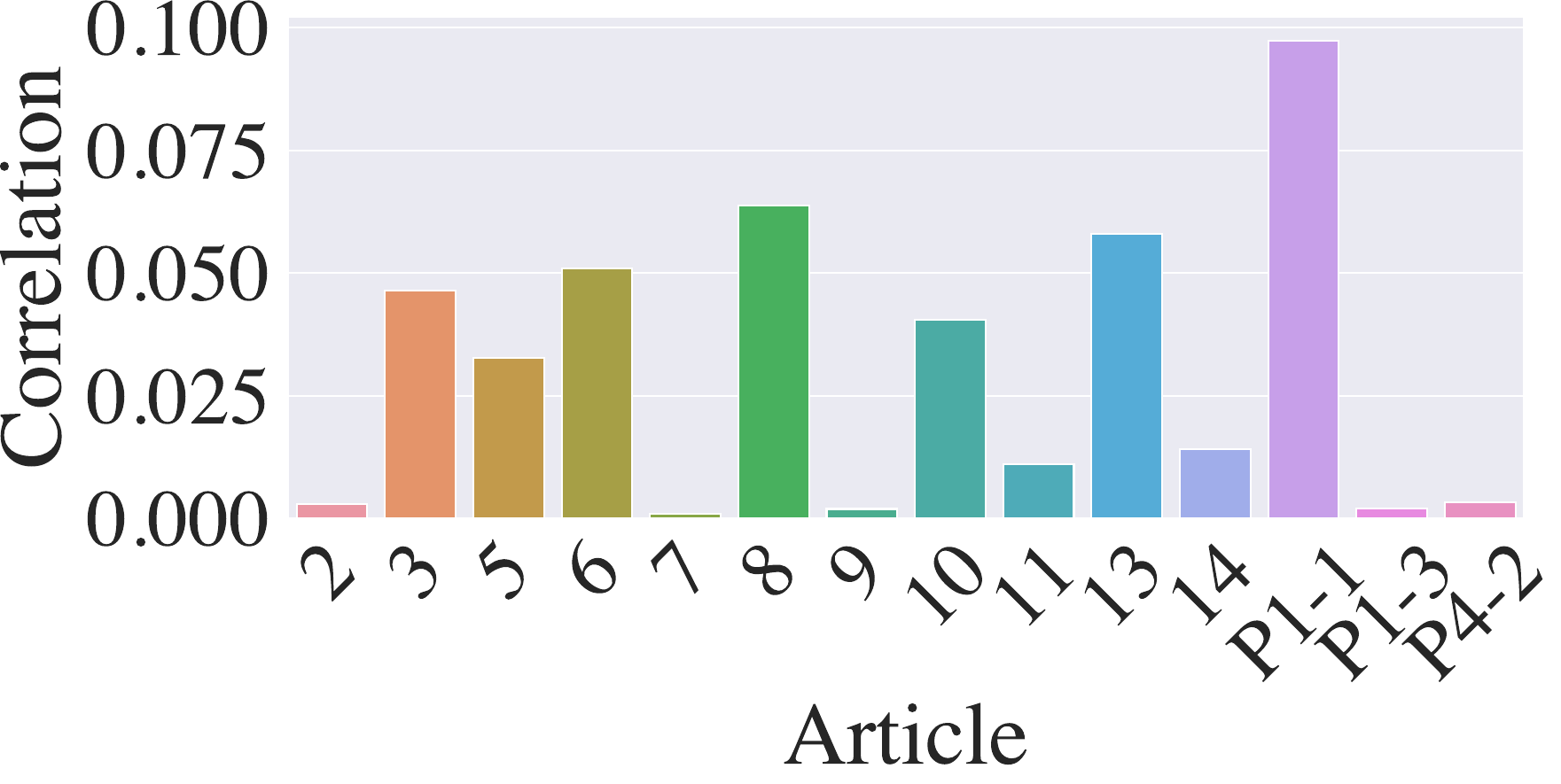} }}\\
    \subfloat[Applied-Negative]{{\includegraphics[width=7.5cm]{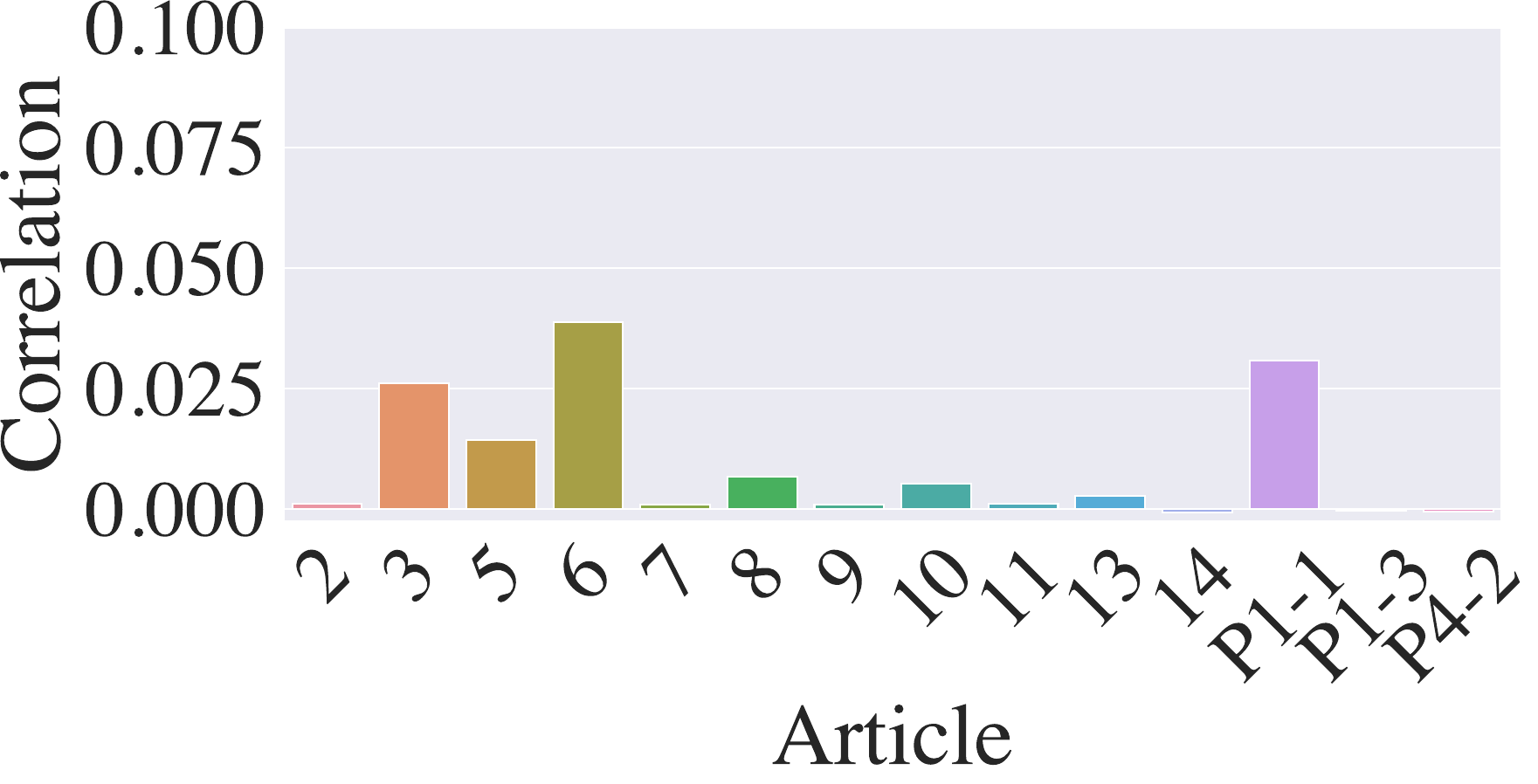} }}\\
    \subfloat[Distinguished-Positive]{{\includegraphics[width=7.5cm]{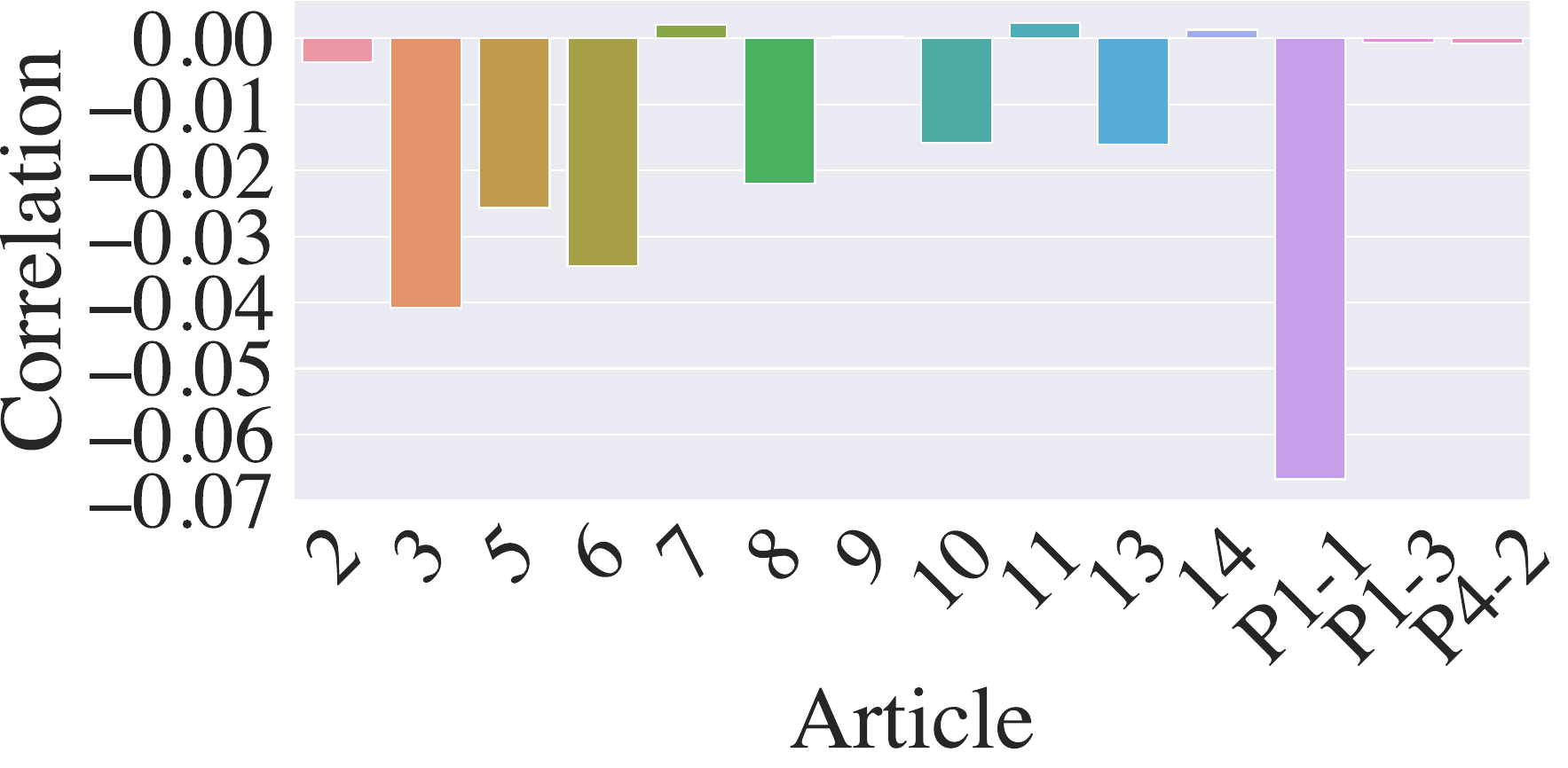} }}\\
    \subfloat[Distinguished-Negative]{{\includegraphics[width=7.5cm]{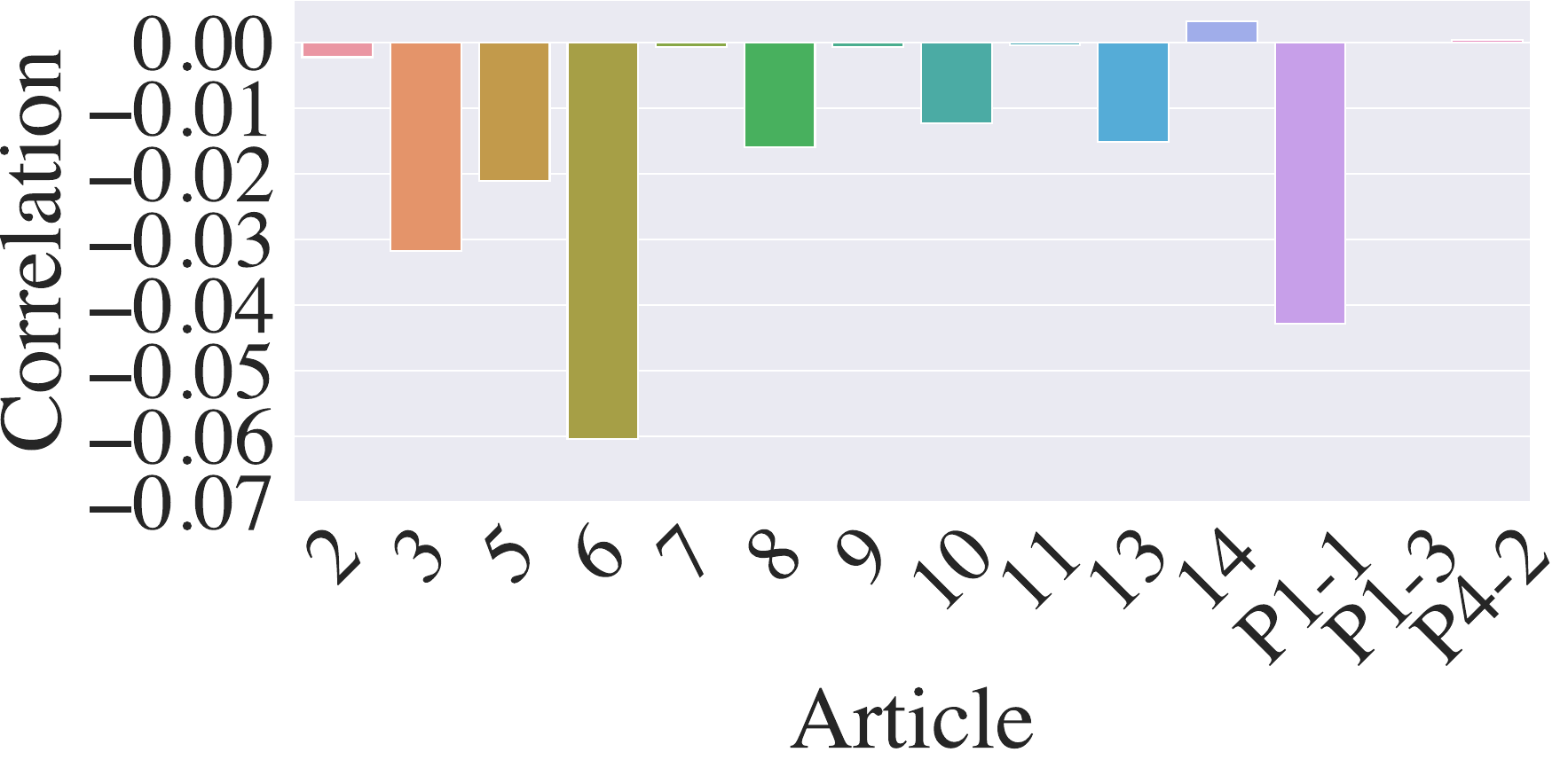} }}\\
    \caption{Spearman's correlation for claimed precedent under the \simple{} with a BERT encoder.}
    \label{fig:article}%
\end{figure}

\newpage

\begin{figure}
    \centering
    \subfloat[Applied-Positive]{{\includegraphics[width=7.5cm]{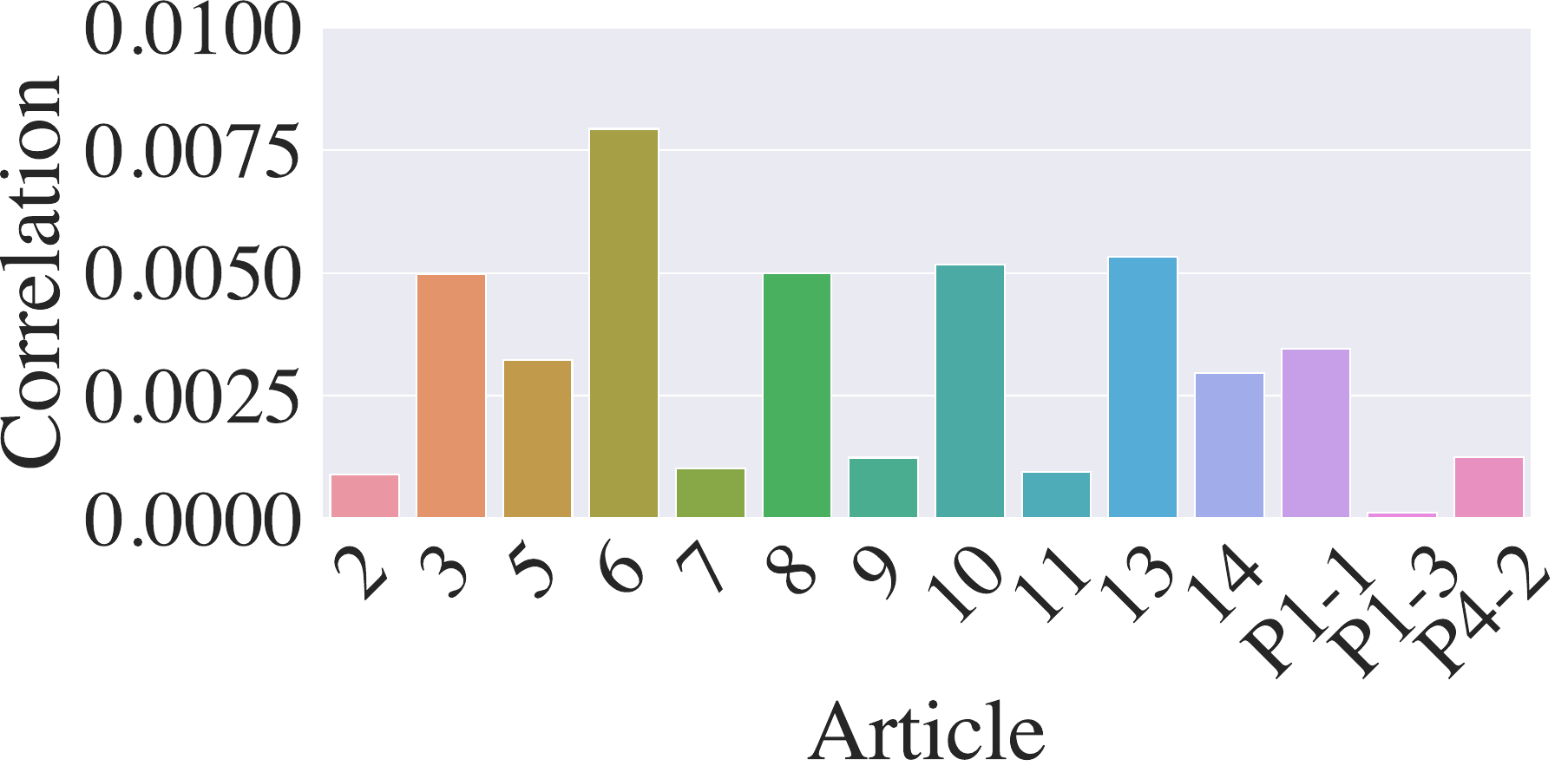} }}\\
    \subfloat[Applied-Negative]{{\includegraphics[width=7.5cm]{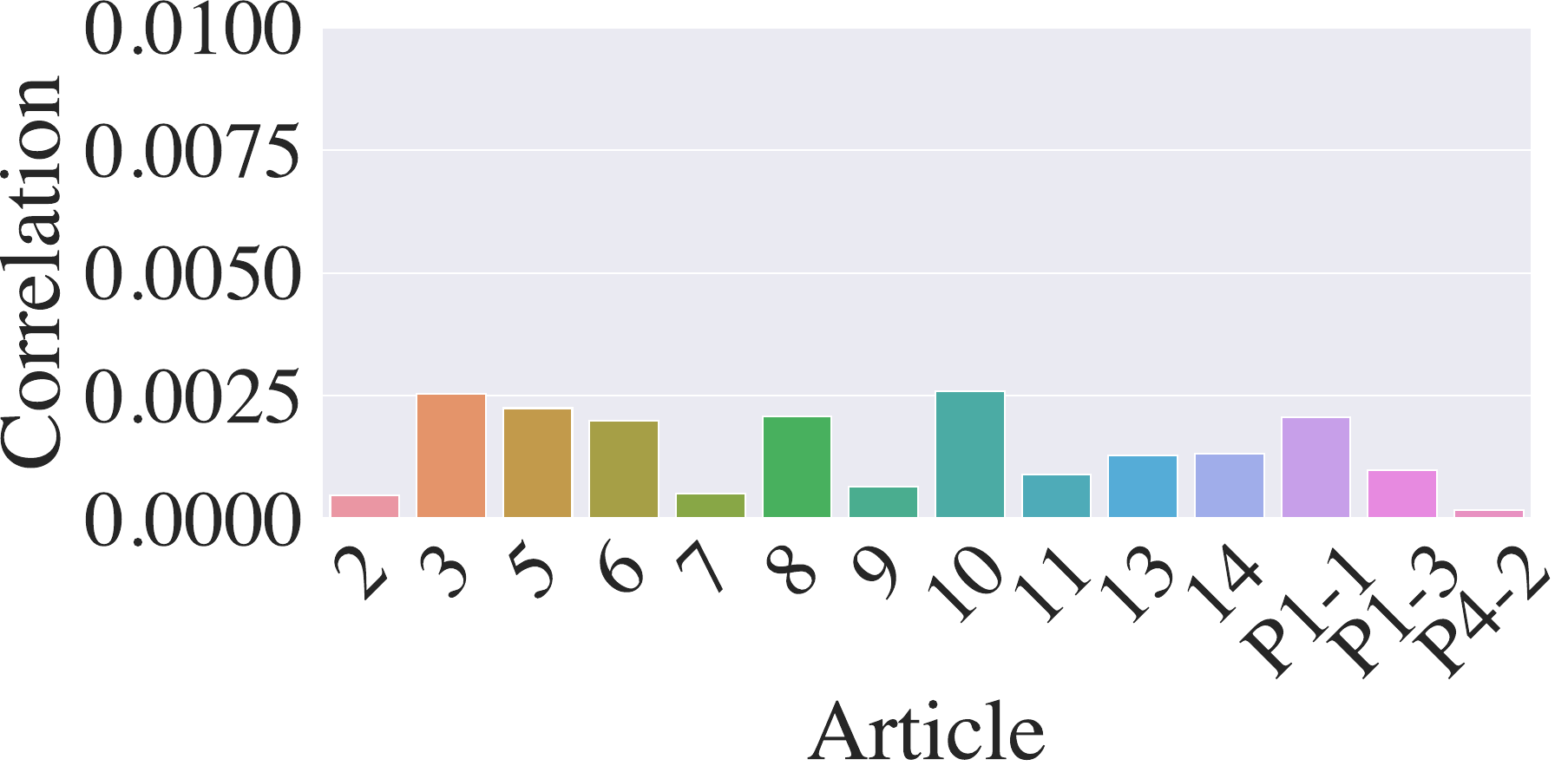} }}\\
    \subfloat[Distinguished-Positive]{{\includegraphics[width=7.5cm]{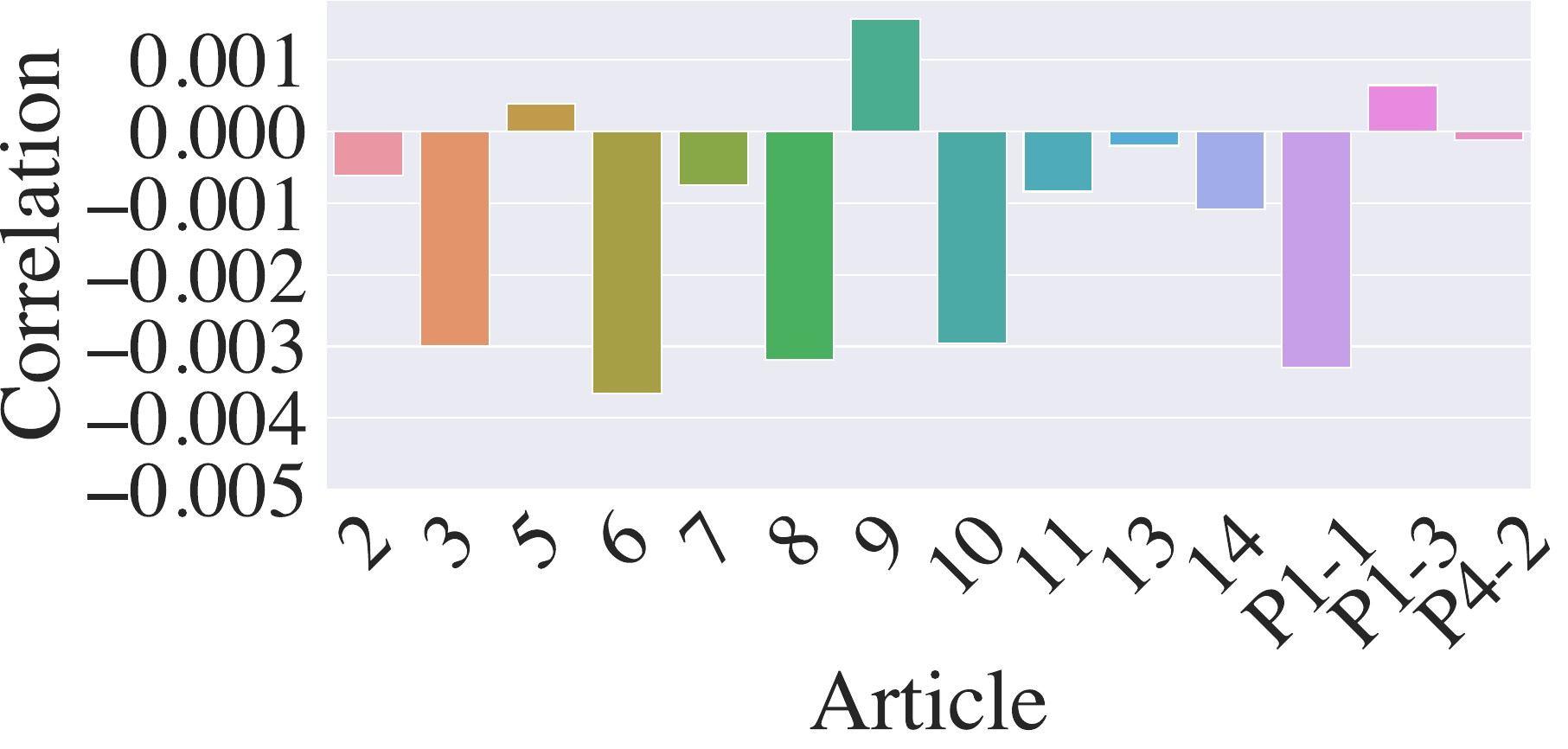} }}\\
    \subfloat[Distinguished-Negative]{{\includegraphics[width=7.5cm]{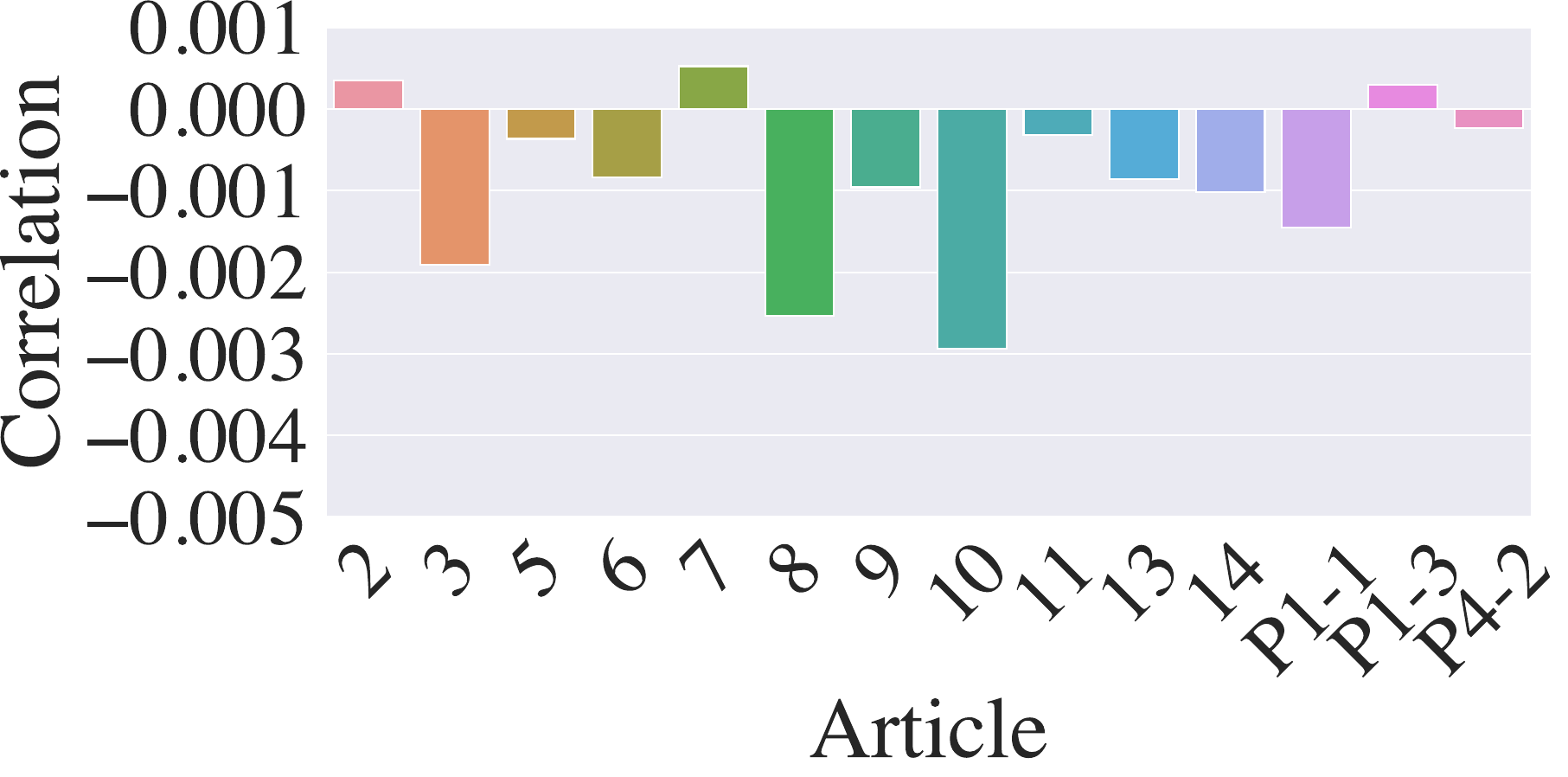} }}\\
    \caption{Spearman's correlation for cited precedent under the \simple{} with a BERT encoder.}
    \label{fig:article}%
\end{figure}

\begin{figure}
    \centering
    \subfloat[Applied-Positive]{{\includegraphics[width=7.5cm]{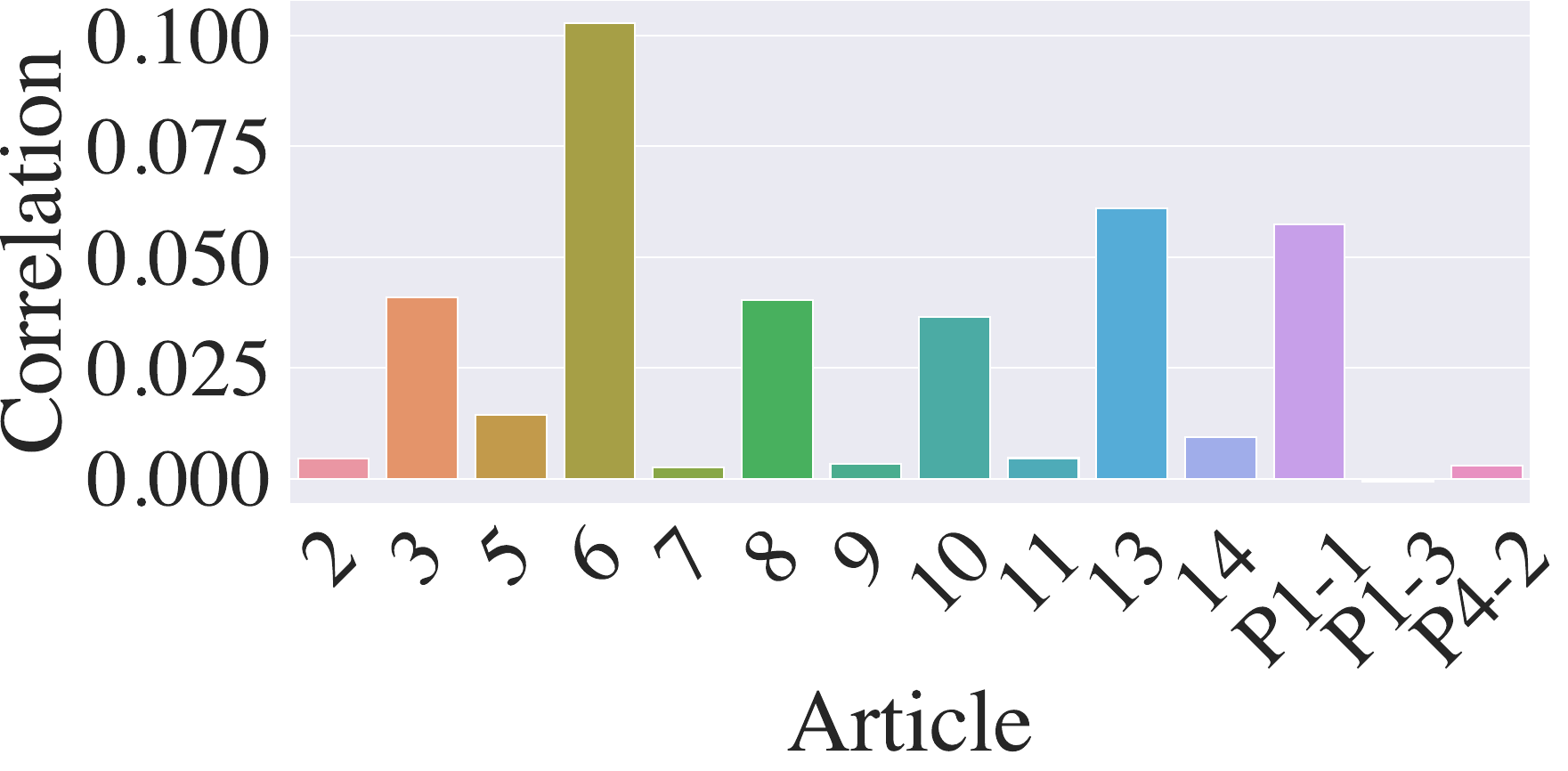} }}\\
    \subfloat[Applied-Negative]{{\includegraphics[width=7.5cm]{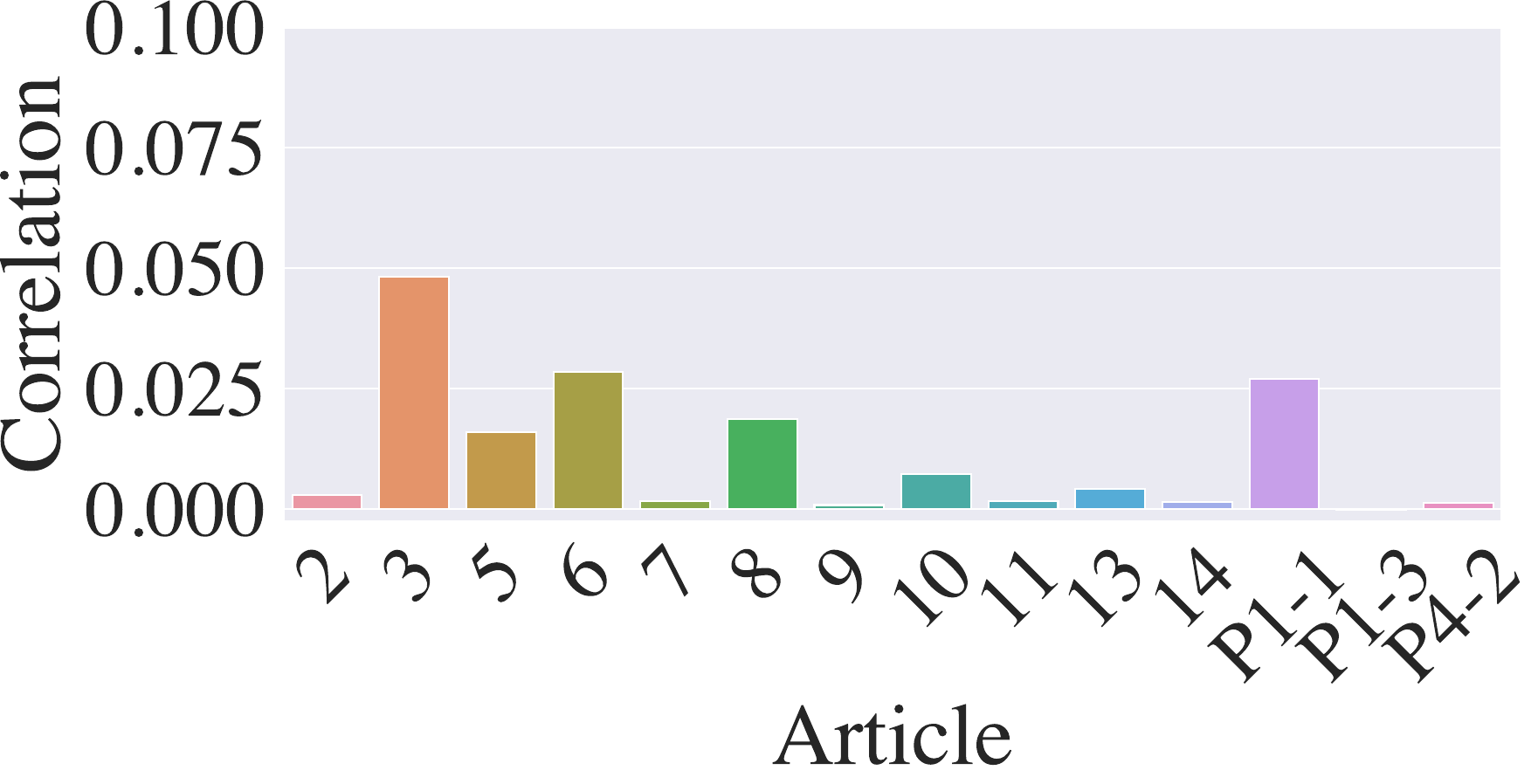} }}\\
    \subfloat[Distinguished-Positive]{{\includegraphics[width=7.5cm]{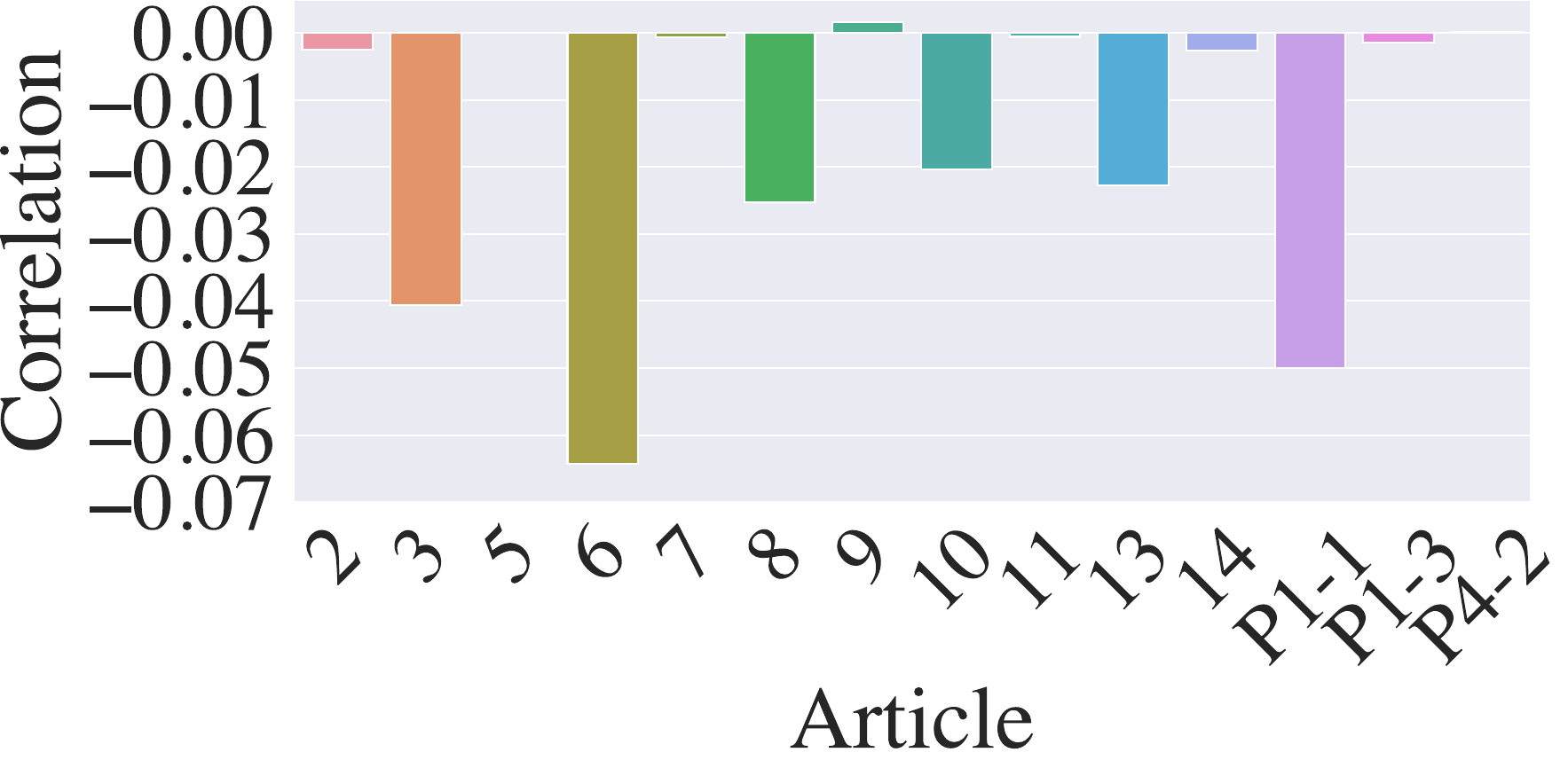} }}\\
    \subfloat[Distinguished-Negative]{{\includegraphics[width=7.5cm]{images/appendix_legal_bert_wide_plot_negative_distinguished.pdf} }}\\
    \caption{Spearman's correlation for all types of precedent under the \simple{} with a LEGAL-BERT encoder.}
    \label{fig:article}%
\end{figure}

\newpage

\begin{figure}
    \centering
    \subfloat[Applied-Positive]{{\includegraphics[width=7.5cm]{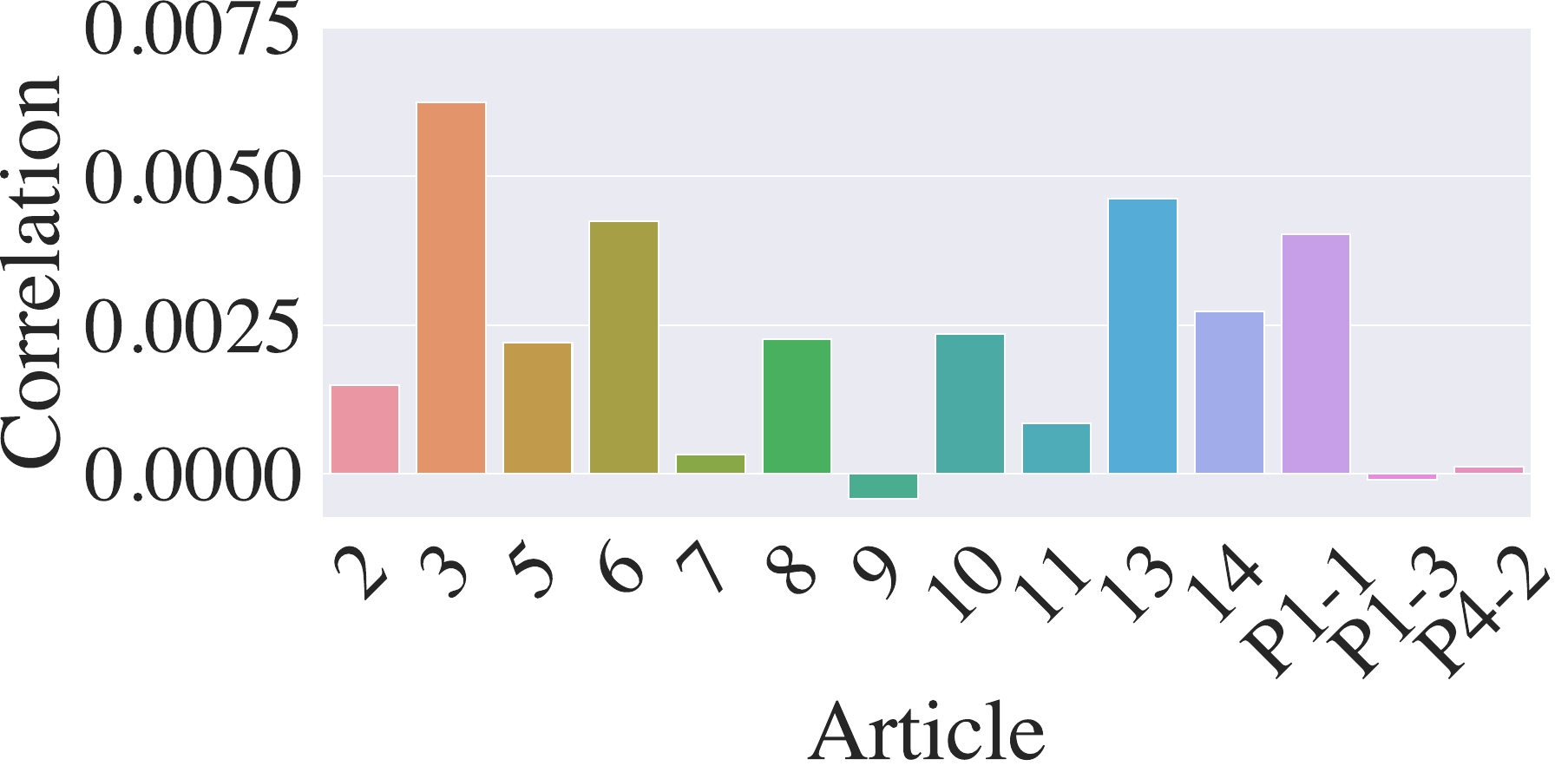} }}\\
    \subfloat[Applied-Negative]{{\includegraphics[width=7.5cm]{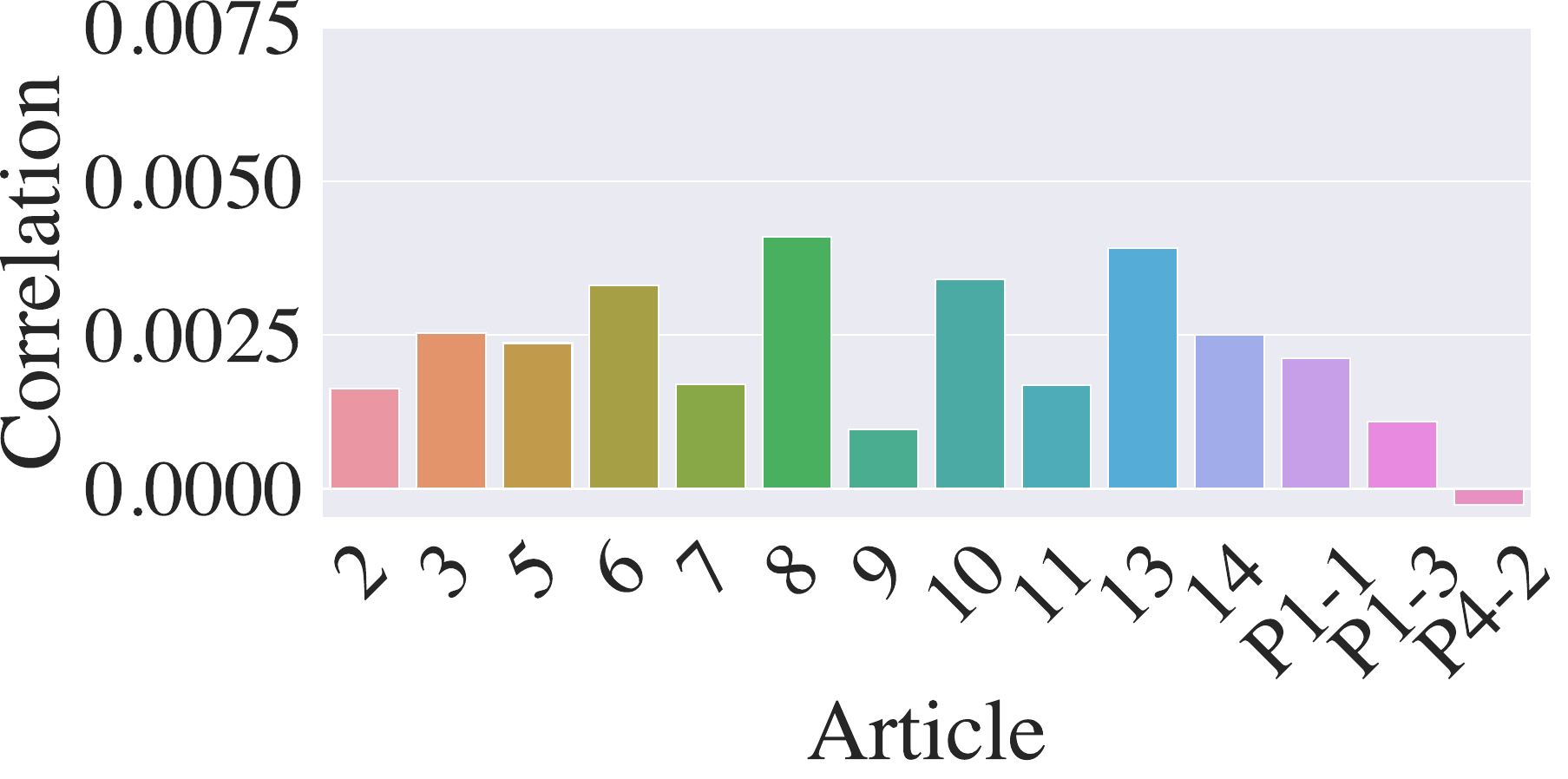} }}\\
    \subfloat[Distinguished-Positive]{{\includegraphics[width=7.5cm]{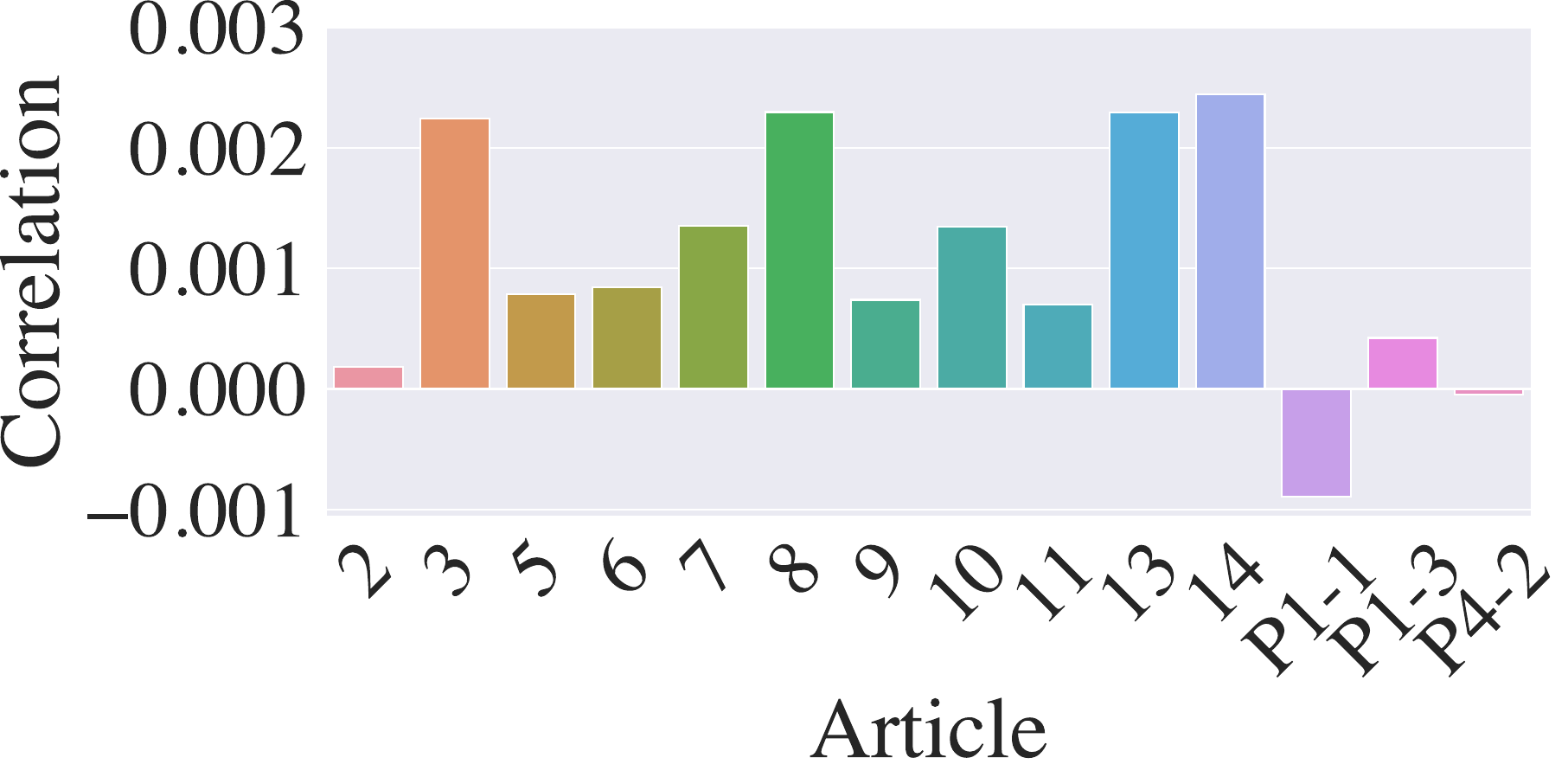} }}\\
    \subfloat[Distinguished-Negative]{{\includegraphics[width=7.5cm]{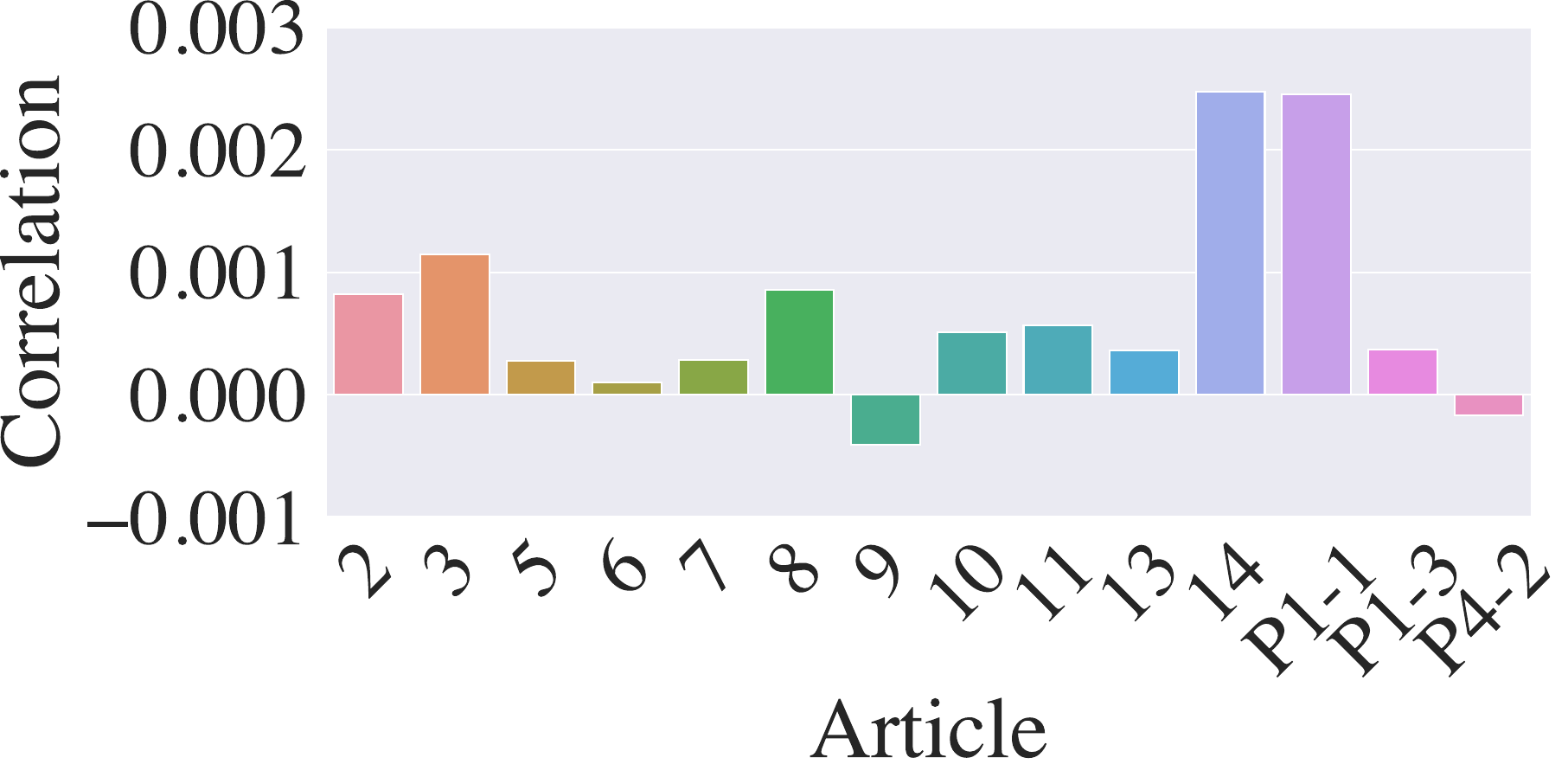} }}\\
    \caption{Spearman's correlation for cited precedent under the \joint{} with a BERT encoder.}
    \label{fig:article}%
\end{figure}

\begin{figure}
    \centering
    \subfloat[Applied-Positive]{{\includegraphics[width=7.5cm]{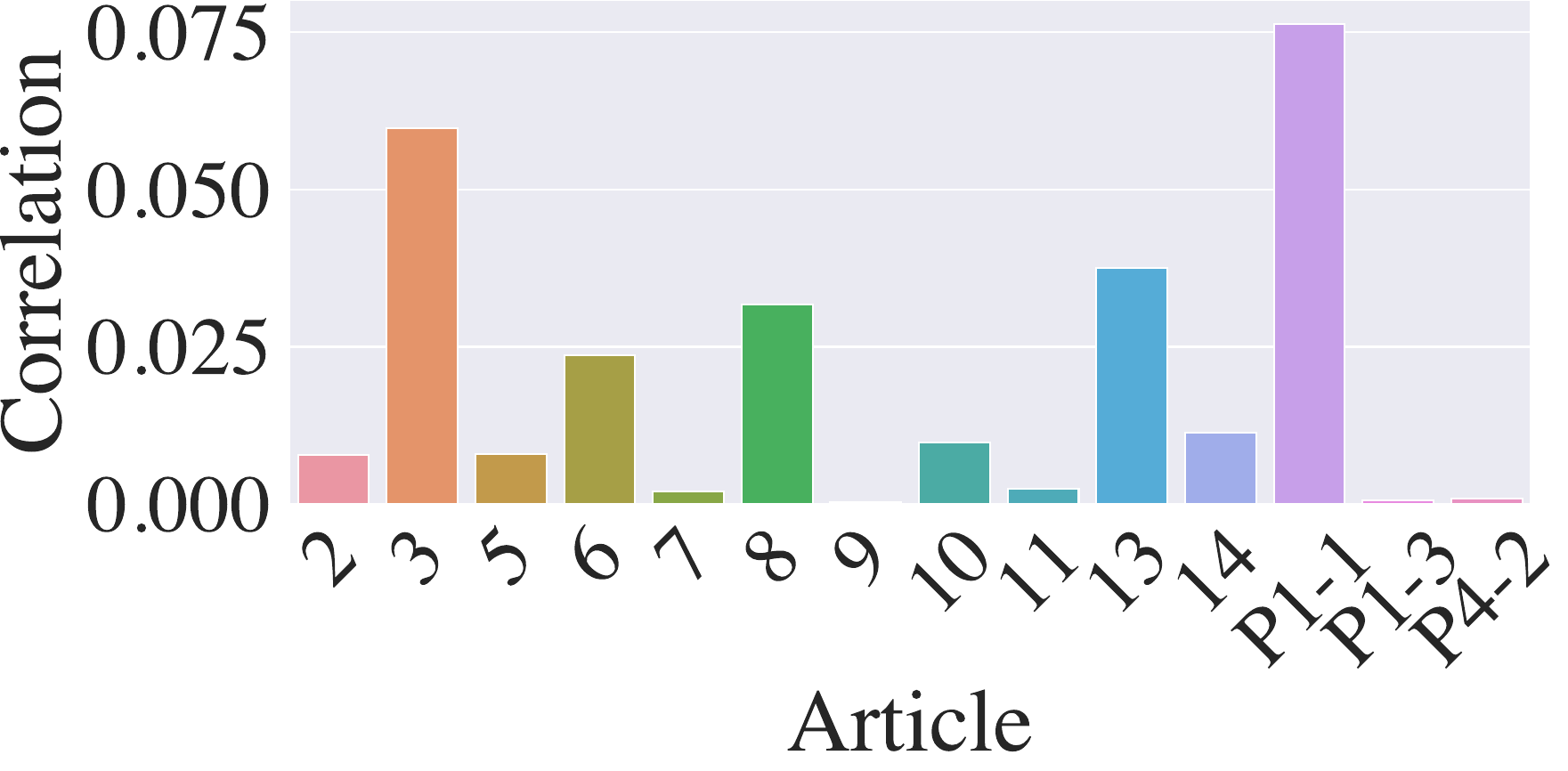} }}\\
    \subfloat[Applied-Negative]{{\includegraphics[width=7.5cm]{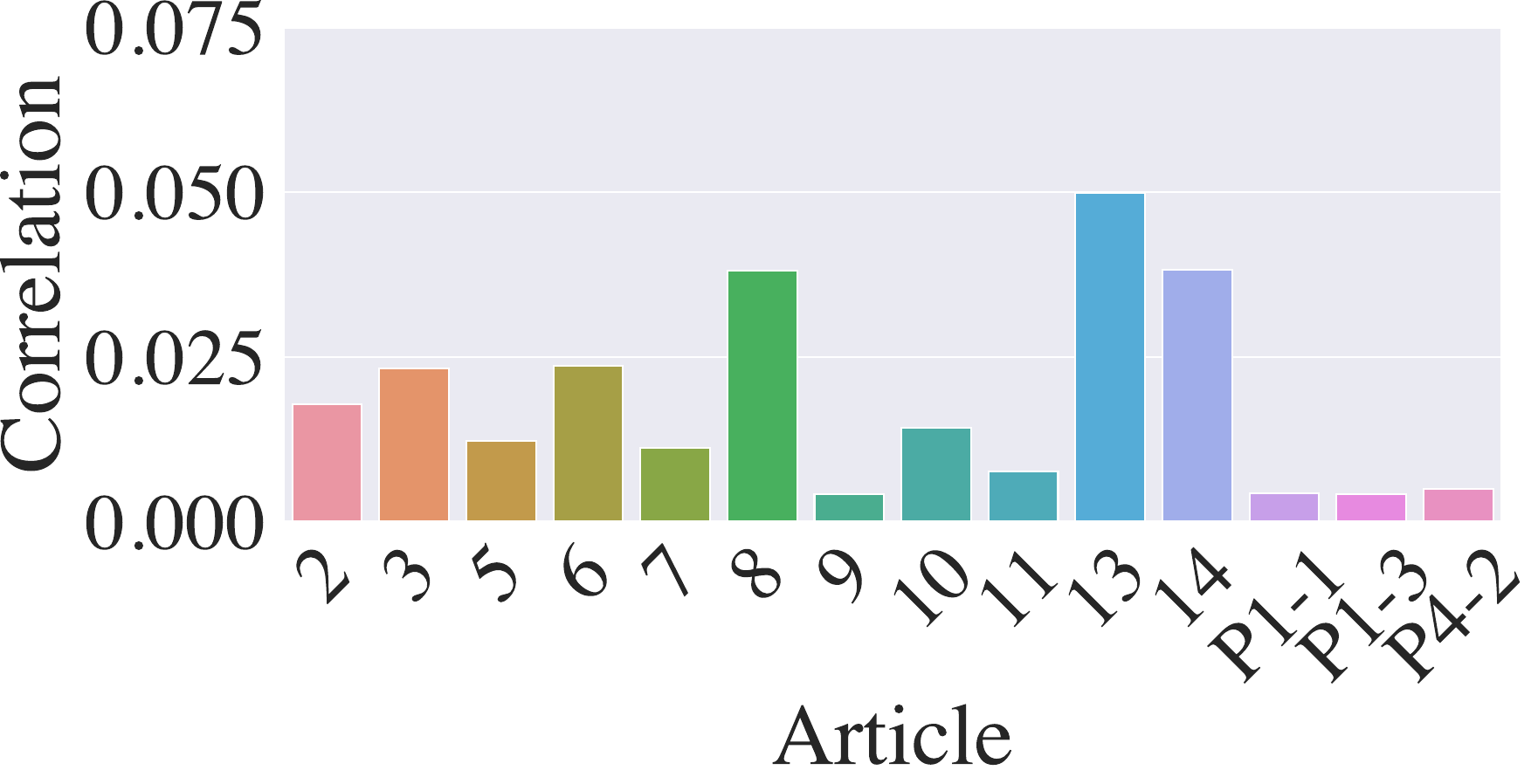} }}\\
    \subfloat[Distinguished-Positive]{{\includegraphics[width=7.5cm]{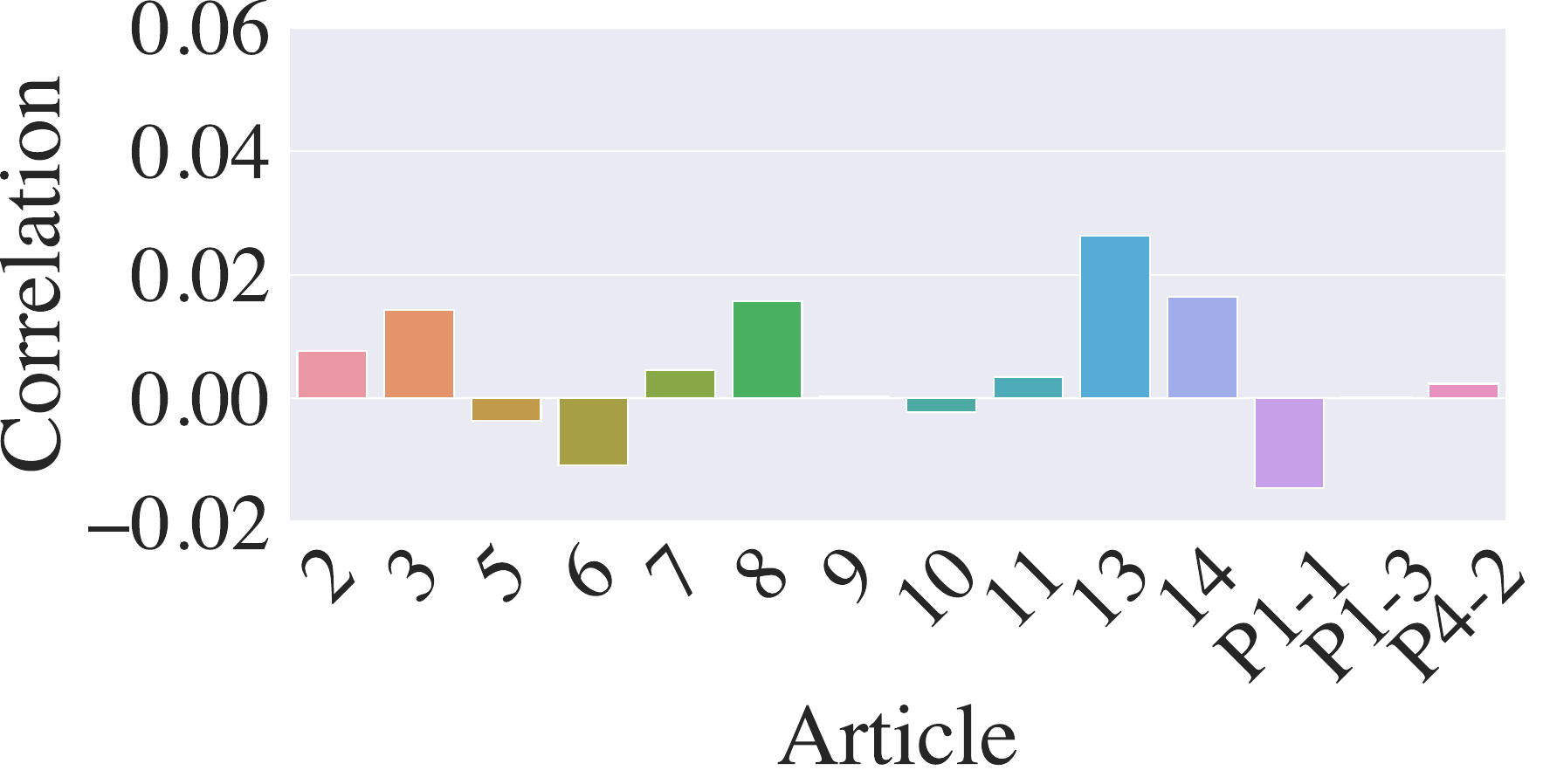} }}\\
    \subfloat[Distinguished-Negative]{{\includegraphics[width=7.5cm]{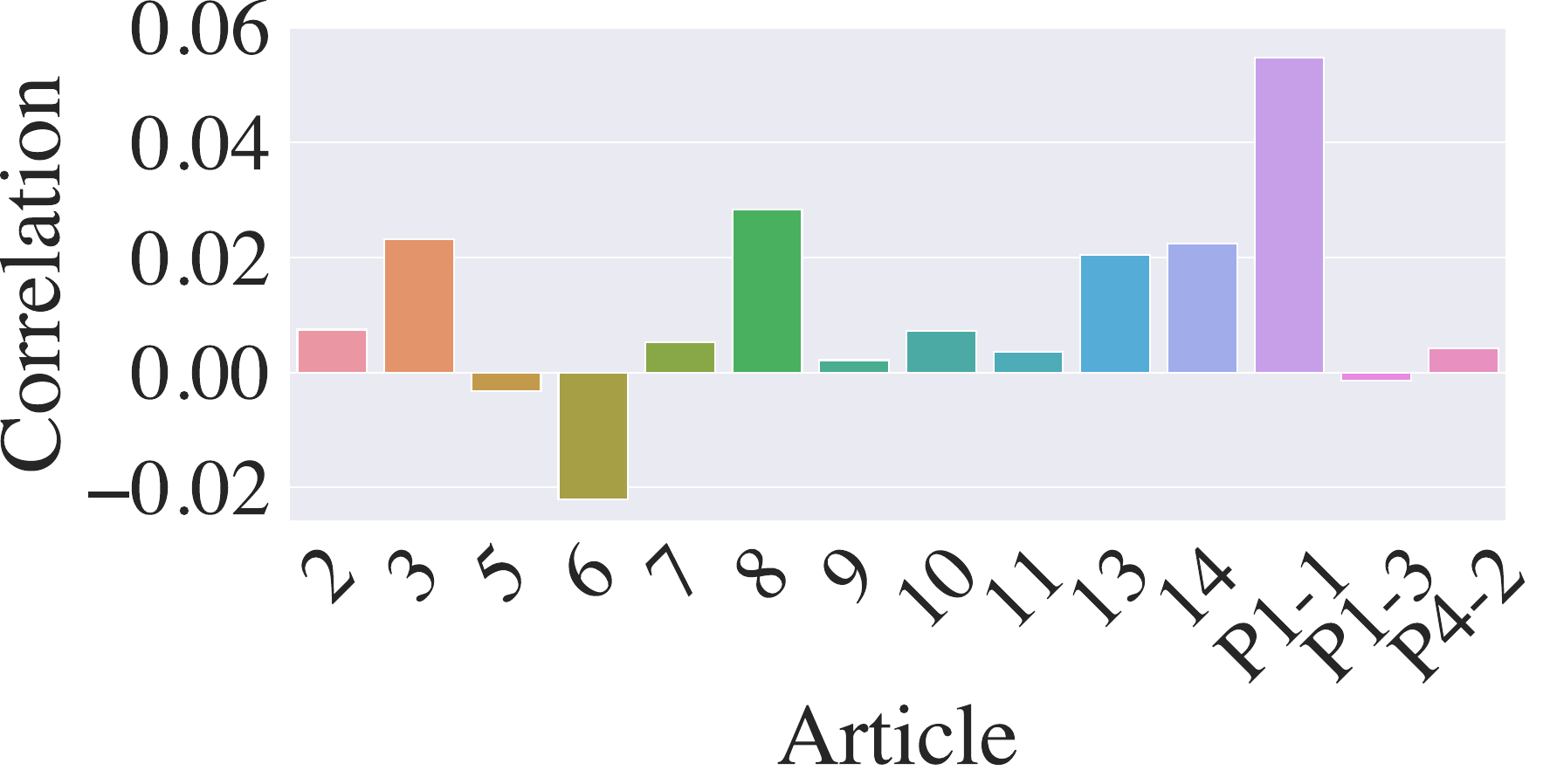} }}\\
    \caption{Spearman's correlation for claimed precedent under the \joint{} with a BERT encoder.}
    \label{fig:article}%
\end{figure}

\begin{figure}
    \centering
    \subfloat[Applied-Positive]{{\includegraphics[width=7.5cm]{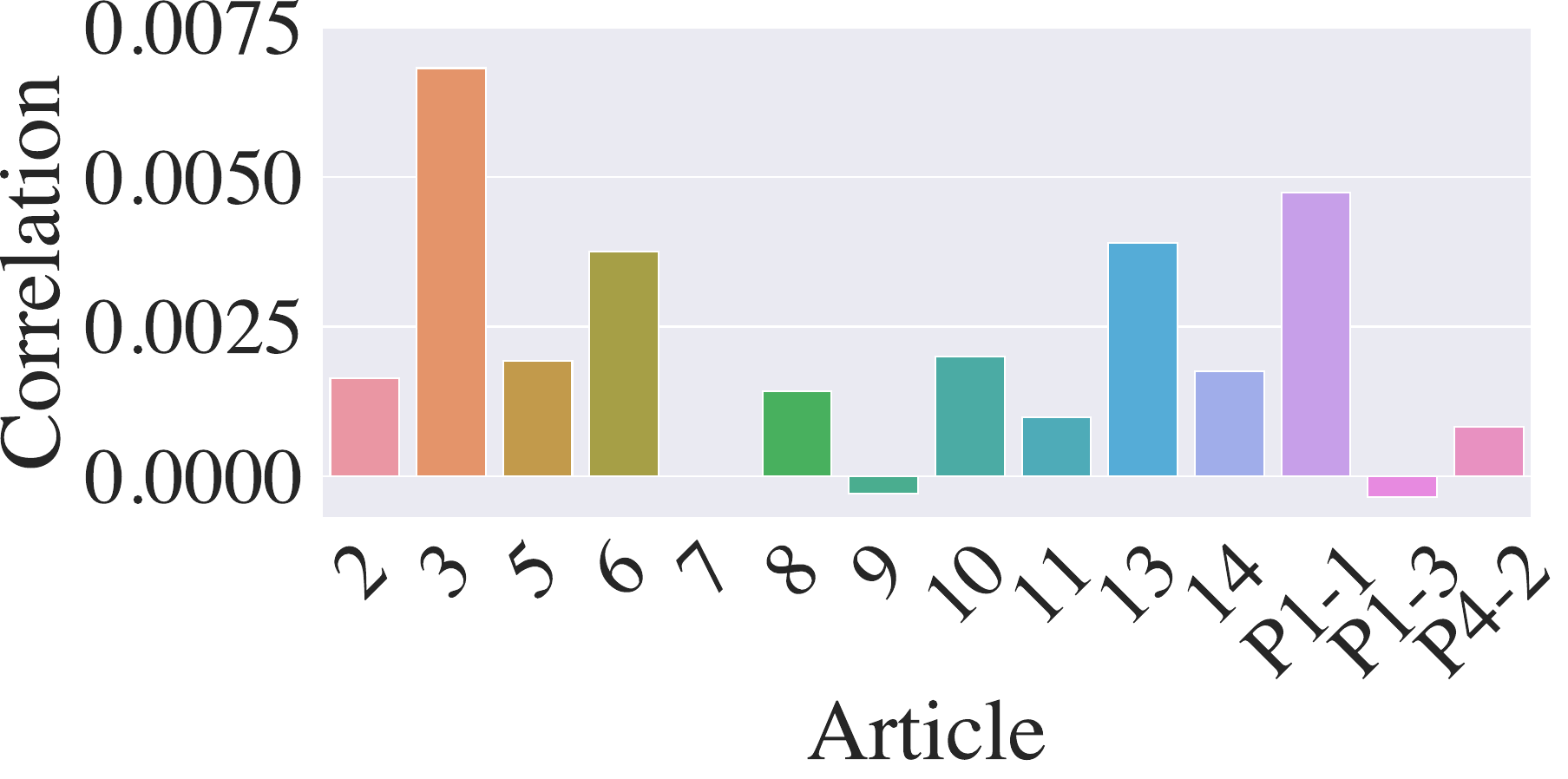} }}\\
    \subfloat[Applied-Negative]{{\includegraphics[width=7.5cm]{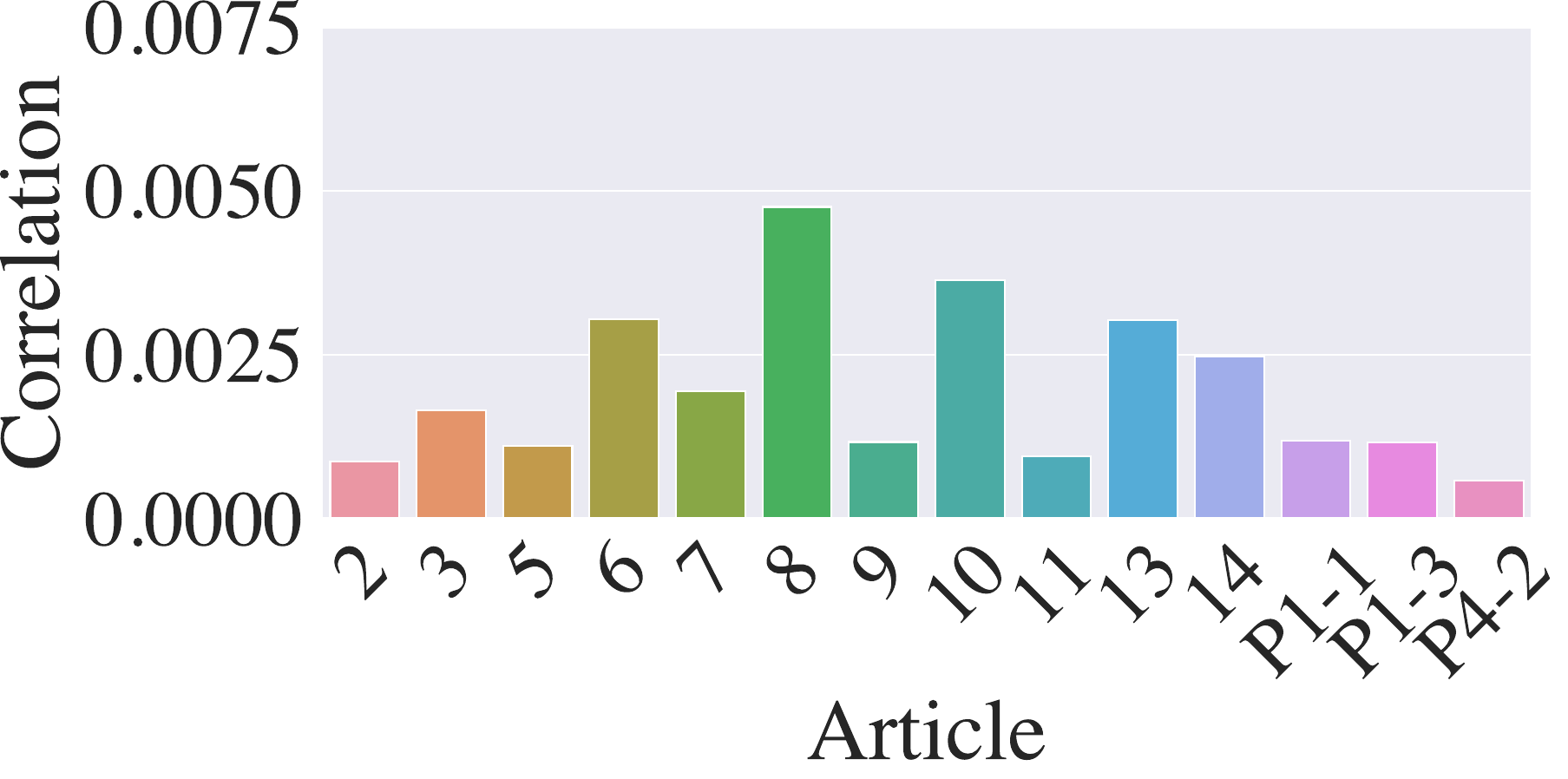} }}\\
    \subfloat[Distinguished-Positive]{{\includegraphics[width=7.5cm]{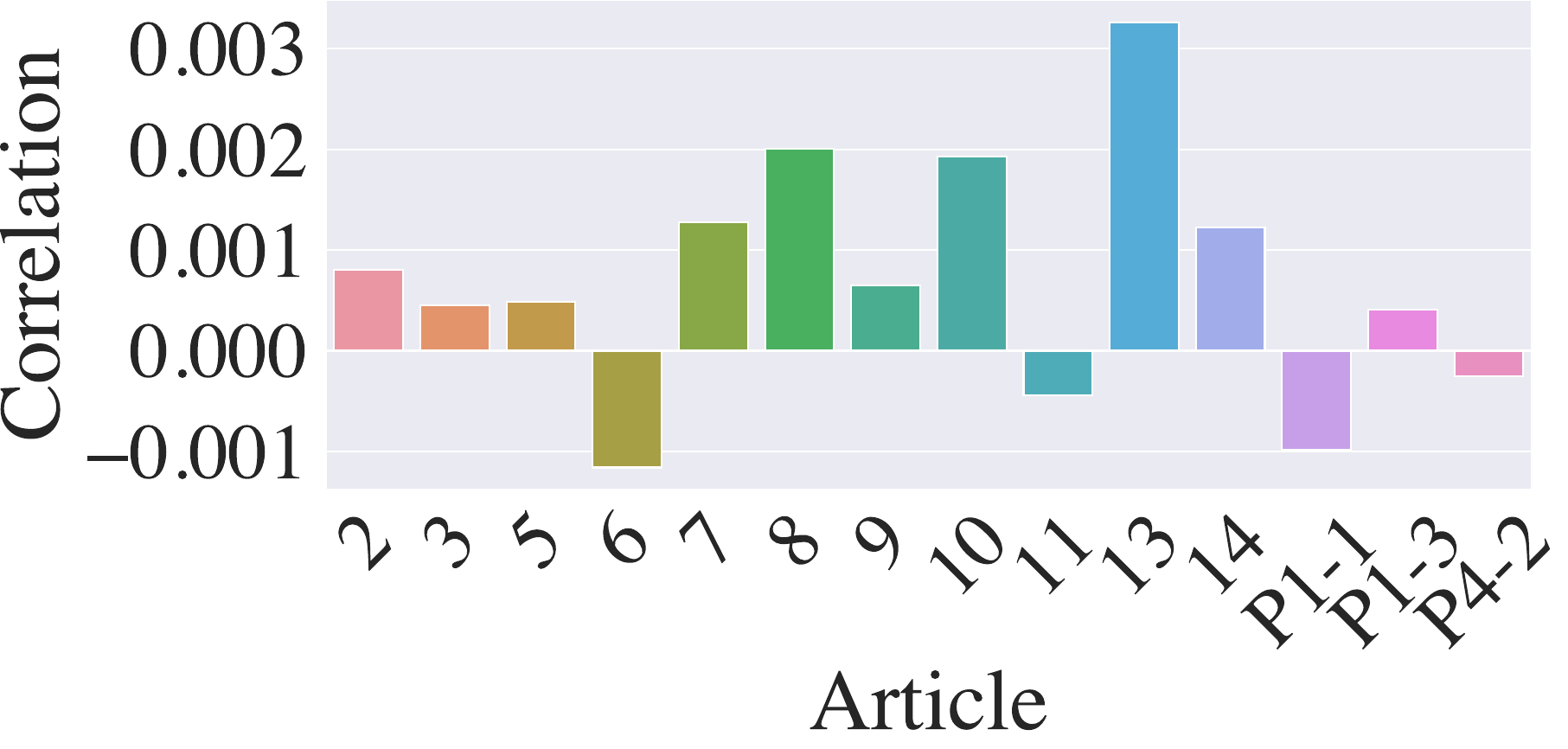} }}\\
    \subfloat[Distinguished-Negative]{{\includegraphics[width=7.5cm]{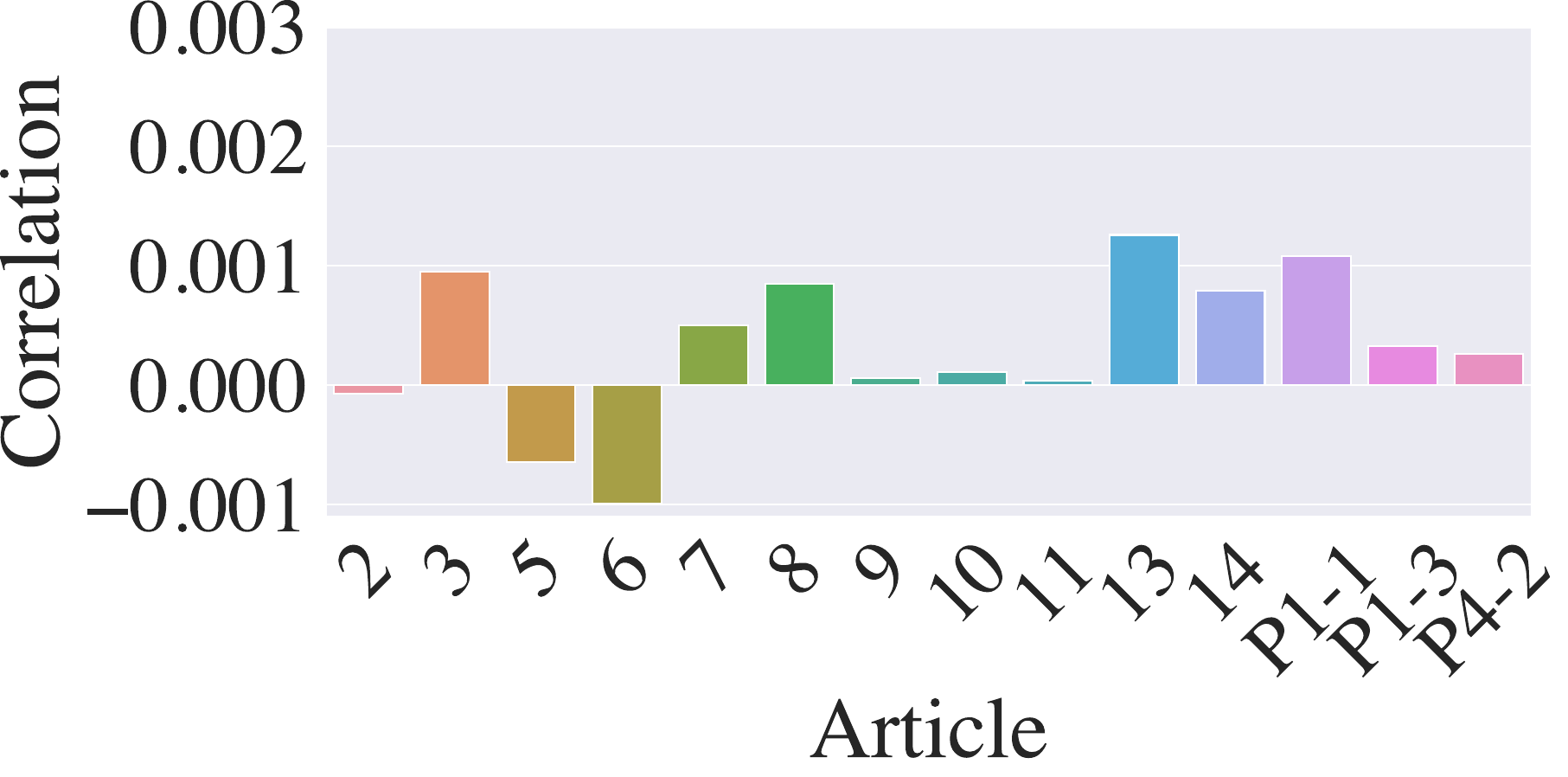} }}\\
    \caption{Spearman's correlation for cited precedent under the \joint{} with a LEGAL-BERT encoder.}
    \label{fig:article}%
\end{figure}

\begin{figure}
    \centering
    \subfloat[Applied-Positive]{{\includegraphics[width=7.5cm]{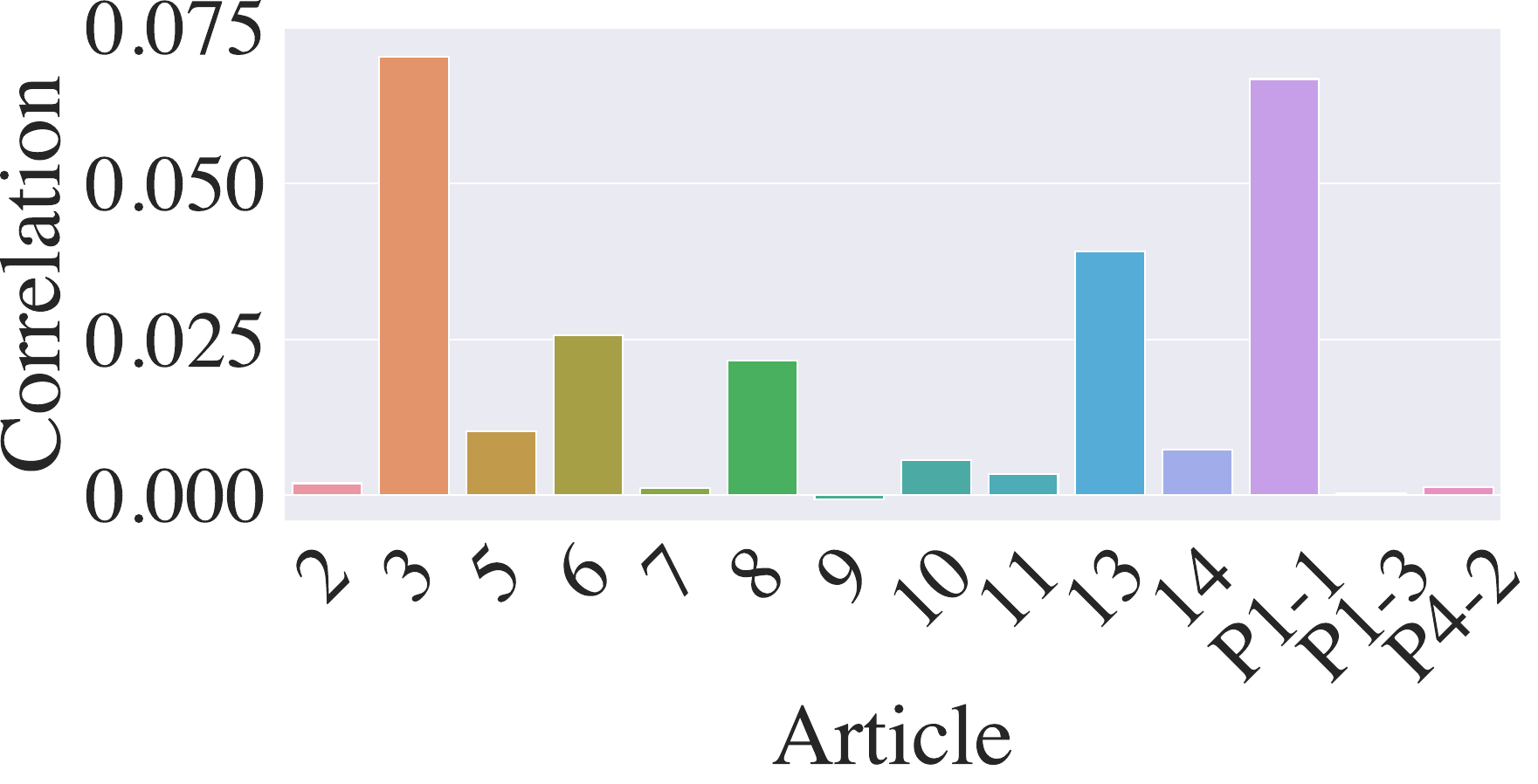} }}\\
    \subfloat[Applied-Negative]{{\includegraphics[width=7.5cm]{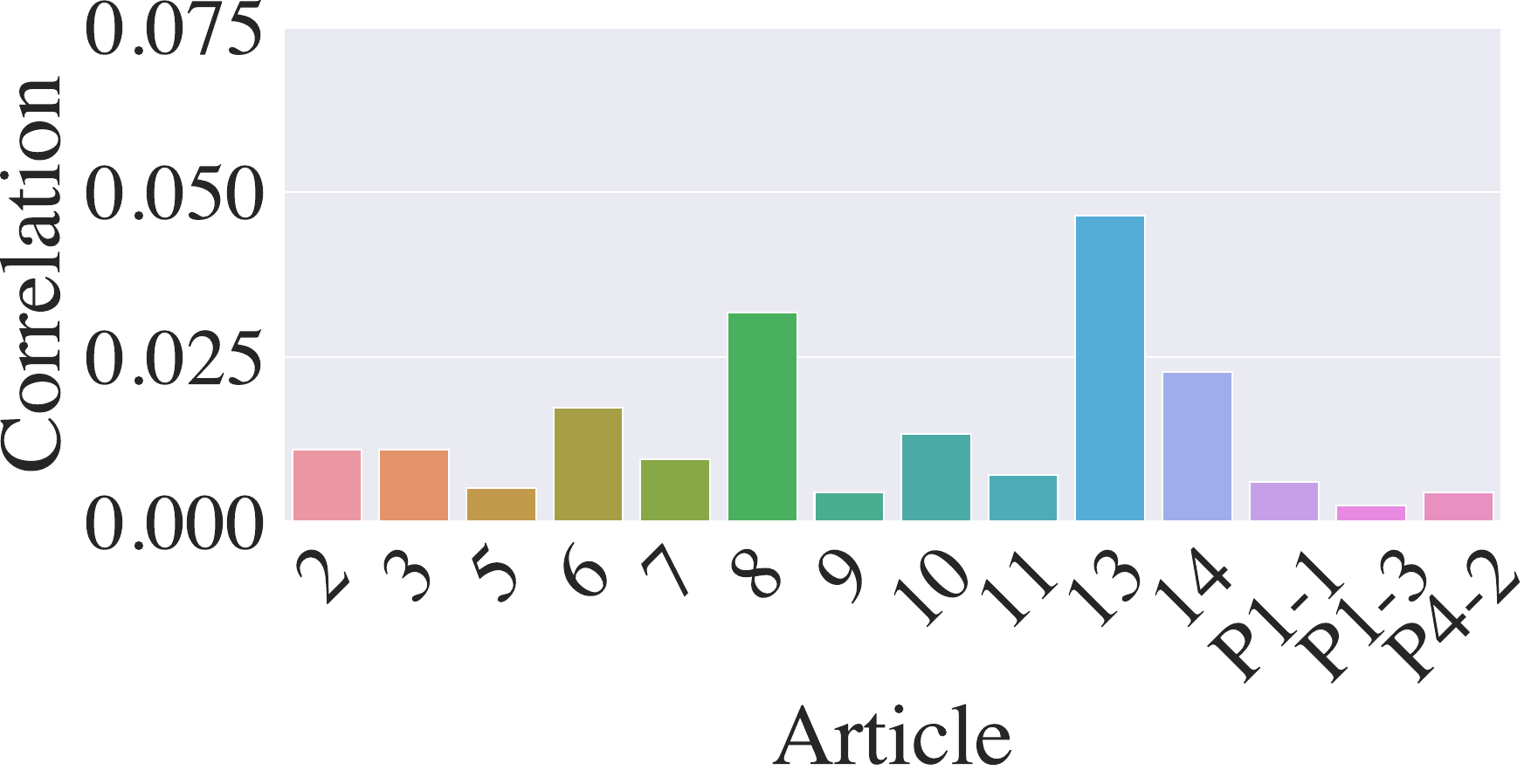} }}\\
    \subfloat[Distinguished-Positive]{{\includegraphics[width=7.5cm]{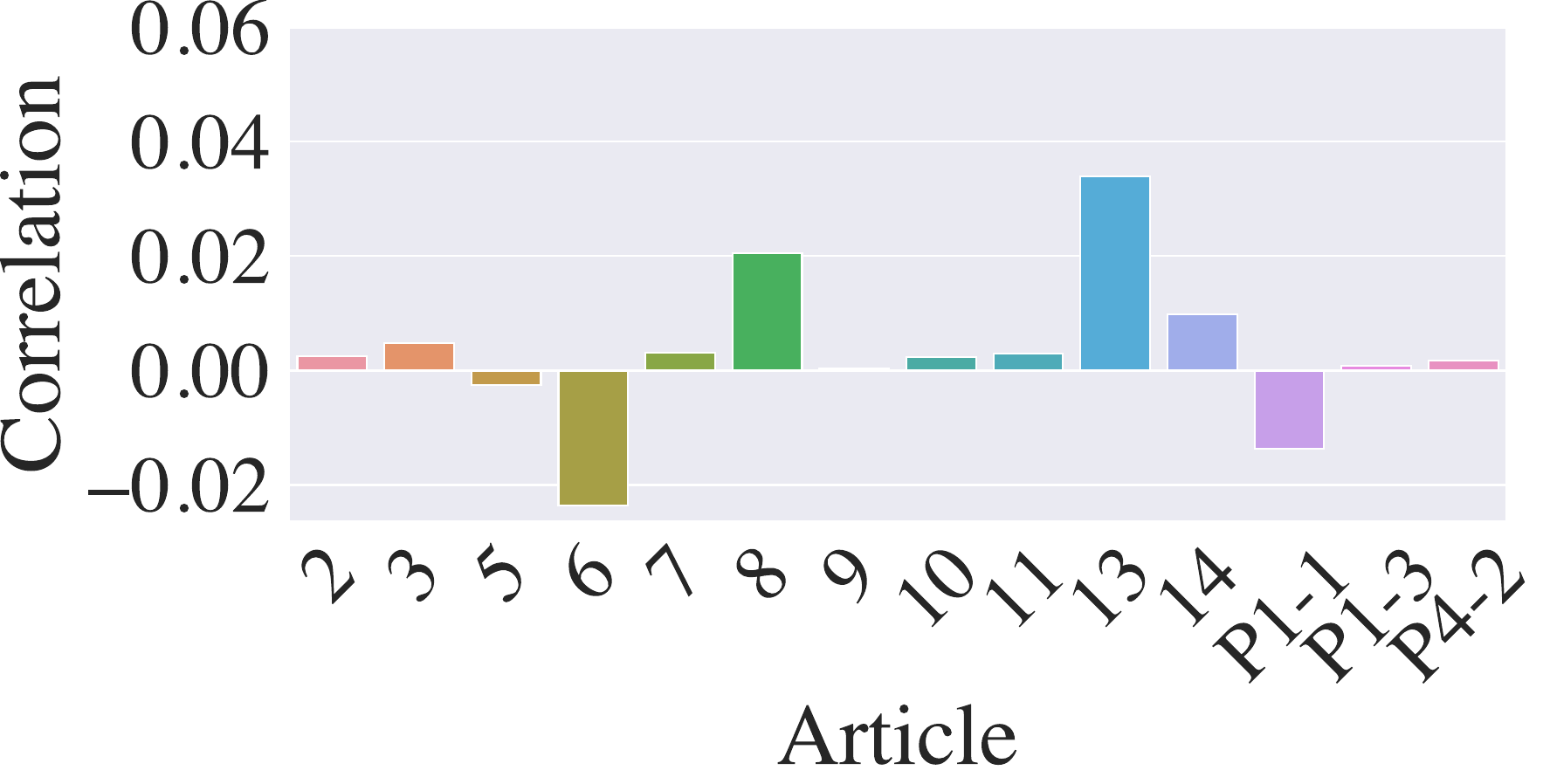} }}\\
    \subfloat[Distinguished-Negative]{{\includegraphics[width=7.5cm]{images/appendix_joint_legal_bert_wide_plot_negative_distinguished.pdf} }}\\
    \caption{Spearman's correlation for claimed precedent under the \joint{} with a LEGAL-BERT encoder.}
    \label{fig:article}%
\end{figure}

\onecolumn
\section{Derivation of the Influence Score}\label{app:alsoinfluence}
We derive \Cref{eq:influence} below.
Consider the following bivariate function of $\btheta$ and $\varepsilon$
\begin{equation}
    F(\btheta, \varepsilon) \defeq \derivative_{\btheta}\Risk(\btheta) + \strength \regularizer(\btheta) + \varepsilon \Loss_{\z}(\btheta) = 0 
\end{equation}
Under suitable regularity conditions on $F$ \citep[Chapter 9]{rudin1976principles}, the implicit function theorem guarantees us the following:
\begin{enumerate}
    \item There exists a function $\btheta^\star_z(\varepsilon)$ of $\varepsilon$ such that $F(\btheta^\star_z(\varepsilon), \varepsilon) = 0$ in a certain neighborhood;
    \item $\btheta^\star_z(\varepsilon)$ is once differentiable with respect to $\varepsilon$ on that neighborhood.
\end{enumerate}
See \citet[Chapter~9]{rudin1976principles} for more detailed treatment.
We now seek to derive $\frac{\dev \btheta^\star_z(\varepsilon)}{\dev \varepsilon}$.
We start with the linearity of the gradient
\begin{equation}
    0 = \derivative_{\btheta}(\Risk(\params) + \strength \regularizer(\params) + \varepsilon \Loss_{\z}(\params)) = \derivative_{\btheta} \Risk(\params) + \strength     \derivative_{\btheta}\regularizer(\params) + \varepsilon   \derivative_{\btheta} \Loss_{\z}(\params) 
\end{equation}
Consider the following manipulation based on taking a derivative with respect to $\varepsilon$:
\begin{subequations}
\begin{align}
    %
&0 =  \frac{\dev 0}{\dev \varepsilon}   = \frac{\dev\left( \derivative_{\btheta} \Risk(\params) + \strength     \derivative_{\btheta}\regularizer(\params) + \varepsilon   \derivative_{\btheta} \Loss_{\z}(\params)\right) }{\dev \varepsilon} &
  \\
&= \frac{\dev  \derivative_{\btheta} \Risk(\params)}{\dev \varepsilon} + \strength     \frac{\dev \derivative_{\btheta}\regularizer(\params)} {\dev \varepsilon}+ \frac{\dev \varepsilon   \derivative_{\btheta} \Loss_{\z}(\params) }{\dev \varepsilon} 
    \\
&= \frac{\dev  \derivative_{\btheta} \Risk(\params)}{\dev \varepsilon} + \strength     \frac{\dev \derivative_{\btheta}\regularizer(\params)} {\dev \varepsilon}+ \derivative_{\btheta} \Loss(\z)(\params) 
 + \varepsilon \frac{\dev \derivative_{\btheta} \Loss_{\z}(\params) }{\dev \varepsilon} 
    \\
&= \derivative_{\btheta}^2 \Risk(\params)\frac{\dev  \params} {\dev \varepsilon} + \strength     \derivative^2_{\btheta}\regularizer(\params) \frac{\dev  \params} {\dev \varepsilon}+ \derivative_{\btheta} \Loss_{\z}(\params) 
 + \varepsilon  \derivative_{\btheta}^2 \Loss_{\z}(\params) \frac{\dev \params}{\dev \varepsilon}
\end{align}
\end{subequations}
\noindent Now, we have
\begin{subequations}
\begin{align}
-\derivative_{\btheta} \Loss_{\z}(\params) 
 &= \derivative_{\btheta}^2 \Risk(\params)\frac{\dev  \btheta} {\dev \varepsilon} + \strength     \derivative^2_{\btheta}\regularizer(\params) \frac{\dev  \btheta} {\dev \varepsilon}
 + \varepsilon  \derivative_{\btheta}^2 \Loss_{\z}(\params) \frac{\dev \btheta}{\dev \varepsilon} \\
  &= \left(\derivative_{\btheta}^2 \Risk(\params) + \strength     \derivative^2_{\btheta}\regularizer(\params) 
 + \varepsilon  \derivative_{\btheta}^2 \Loss_{\z}(\params)  \right)\frac{\dev \btheta}{\dev \varepsilon}
 \end{align}
 \end{subequations}
 Evaluating the expression at $\varepsilon = 0$, we arrive at
 \begin{equation}
     \frac{\dev \params}{\dev \varepsilon}\Big|_{\varepsilon = 0} = - \left(\derivative_{\btheta}^2 \Risk(\btheta^\star) + \strength     \derivative^2_{\btheta}\regularizer(\btheta^\star) \right)^{-1}\derivative_{\btheta} \Loss_{\z}(\btheta^\star) 
 \end{equation}
 where $\btheta^\star$ are the parameters obtained in \Cref{eq:optimization-problem}.

\newpage
\section{Additional Results}\label{app:modelbased}

Here are the full results for our analysis in \cref{sec:alternativeview}.

\begin{table}[hb]
\centering
\begin{tabular}{llllrrrrr}
\toprule
&                    & &                   & \multicolumn{2}{c}{\textbf{Applied}}                                  & \multicolumn{2}{c}{\textbf{Distinguished}} &    \textbf{Overall}                        \\
\midrule
\textbf{Type} & \textbf{Language Model}                    & $\mathbf{F_1}$ &                   & \multicolumn{1}{c}{\textbf{Cited}} & \multicolumn{1}{c}{\textbf{Claimed}} & \multicolumn{1}{c}{\textbf{Cited}} & \multicolumn{1}{c}{\textbf{Claimed}} &  \\
\midrule

Simple & BERT  & 0.64 & \textbf{Positive} & 0.013 & 0.156 & -0.001 & 0.006 & 0.149       \\
                                    & & & \textbf{Negative} & 0.004 & 0.024 & -0.002 & -0.011           \\
        & LEGAL-BERT & 0.67 & \textbf{Positive} & 0.015 & 0.205 & -0.003 & -0.024 & 0.183              \\
                                         & & & \textbf{Negative} & 0.004 & 0.025 & -0.003 & -0.025                        \\       
Joint & BERT & 0.66 & \textbf{Positive} & 0.011 & 0.136 & 0.002 & 0.031 & 0.138        \\
                                        & & & \textbf{Negative} & 0.002 & 0.003 & 0.001 & 0.003                 \\
        & LEGAL-BERT & 0.68 & \textbf{Positive} & 0.009 & 0.146 & 0.002 & 0.017 & 0.148 \\
                                                 & & & \textbf{Negative} & 0.001 & 0.022 & 0.001 & 0.009\\       
   
\bottomrule
\end{tabular}
\caption{Spearman correlations between the influence scores and the precedent for cases where the outcome was correctly predicted.}
\label{table:results_correct}
\end{table}

\begin{table}[hb]
\centering
\begin{tabular}{llllrrrrr}
\toprule
&                    & &                   & \multicolumn{2}{c}{\textbf{Applied}}                                  & \multicolumn{2}{c}{\textbf{Distinguished}} &         \textbf{Overall}         \\
\midrule
\textbf{Type} & \textbf{Language Model}                    & $\mathbf{F_1}$ &                   & \multicolumn{1}{c}{\textbf{Cited}} & \multicolumn{1}{c}{\textbf{Claimed}} & \multicolumn{1}{c}{\textbf{Cited}} & \multicolumn{1}{c}{\textbf{Claimed}} &  \\
\midrule

Simple & BERT  & 0.64 & \textbf{Positive} & 0.002 & 0.004 & 0.004 & 0.042 & 0.007          \\
                                    & & & \textbf{Negative} & 0.000 & -0.027 & 0.001 & 0.009          \\
         & LEGAL-BERT & 0.67 & \textbf{Positive} & 0.005 & 0.029 & 0.003 & 0.038 & 0.032             \\
                                         & & & \textbf{Negative} & 0.000 & -0.011 & 0.000 & 0.000                      \\       
Joint & BERT & 0.66 & \textbf{Positive} & 0.002 & 0.002 & 0.003 & 0.035 & 0.010           \\
                                        & & & \textbf{Negative} & 0.001 & 0.004 & 0.002 & 0.006                 \\
     & LEGAL-BERT & 0.68 & \textbf{Positive} & 0.003 & 0.002 & 0.002 & 0.032 & 0.011 \\
                                                 & & & \textbf{Negative} & 0.000 & 0.008 & 0.002 & 0.013  \\       
   
\bottomrule
\end{tabular}
\caption{Spearman correlations between the influence scores and the precedent for cases where the precedent is model-based.}

\label{table:results_model}
\end{table}

\end{document}